\documentclass[twoside,11pt]{article}

\newcommand{\xv}{\mathbf{x}}

\newcommand{\yv}{\mathbf{y}}

\newcommand{\fv}{\mathbf{f}}

\newcommand{\alphav}{\mathbf{\alpha}}

\newcommand{\wv}{\mathbf{w}}

\newcommand{\xiv}{\mathbf{\xi}}
\newcommand{\tauv}{\mathbf{\tau}}

\newcommand{\ud}{\mathrm{d}}
\def\indicator{{\mathbb I}}
\newcommand{\D}{\scriptscriptstyle}

\usepackage{jmlr2e}
\usepackage{enumerate}
\usepackage{algorithm}
\usepackage{algorithmic}
\usepackage{epsf}
\usepackage{graphicx}
\usepackage{subfigure} 

\begin{document}

\title{Maximum Entropy Discrimination Markov Networks}

\author{\name Jun Zhu \email jun-zhu@mails.tsinghua.edu.cn \\
       \addr State Key Lab of Intelligent Technology $\&$ Systems\\
       \addr Tsinghua National Lab for Information Science and Technology\\
       \addr Department of Computer Science and Technology\\
       Tsinghua University\\
       \AND
       \name Eric P. Xing  \email epxing@cs.cmu.edu\\
       \addr Machine Learning Department\\
       Carnegie Mellon University}

\editor{?}

\maketitle

\begin{abstract}

Standard maximum margin structured prediction methods lack a
straightforward probabilistic interpretation of the learning scheme
and the prediction rule. Therefore its unique advantages such as
dual sparseness and kernel tricks cannot be easily conjoined with
the merits of a probabilistic model such as Bayesian regularization,
model averaging, and allowing hidden variables. In this paper, we
present a novel and general framework called \textit{Maximum Entropy
Discrimination Markov Networks} (MaxEnDNet), which integrates these
two approaches and combines and extends their merits. Major
innovations of this model include: 1) It generalizes the extant
Markov network prediction rule based on a point estimator of weights
to a Bayesian-style estimator that integrates over a learned
distribution of the weights. 2) It extends the conventional
max-entropy discrimination learning of classification rule to a new
structural max-entropy discrimination paradigm of learning the
distribution of Markov networks. 3) It subsumes the well-known and
powerful Maximum Margin Markov network (M$^3$N) as a special case,
and leads to a model similar to an $L_1$-regularized M$^3$N that is
simultaneously primal and dual sparse, or other types of Markov
network by plugging in different prior distributions of the weights.
4) It offers a simple inference algorithm that combines existing
variational inference and convex-optimization based M$^3$N solvers
as subroutines. 5) It offers a PAC-Bayesian style generalization
bound. This work represents the first successful attempt to combine
Bayesian-style learning (based on generative models) with structured
maximum margin learning (based on a discriminative model), and
outperforms a wide array of competing methods for structured
input/output learning on both synthetic and real OCR and web data
extraction data sets.
\end{abstract}


\section{Introduction}

Inferring structured predictions based on high-dimensional, often
multi-modal and hybrid covariates remains a central problem in data
mining (e.g., web-info extraction), machine intelligence (e.g.,
machine translation), and scientific discovery (e.g., genome
annotation). Several recent approaches to this problem are based on
learning discriminative graphical models defined on composite
features that explicitly exploit the structured dependencies among
input elements and structured interpretational outputs. Major
instances of such models include the conditional random fields
(CRFs)~\citep{Lafferty:01}, Markov networks (MNs)~\citep{Taskar:03},
and other specialized graphical models~\citep{Altun:03}. Various
paradigms for training such models based on different loss functions
have been explored, including the maximum conditional likelihood
learning \citep{Lafferty:01} and the max-margin learning
\citep{Altun:03,Taskar:03,Tsochantaridis:04}, with remarkable
success.

The likelihood-based models for structured predictions are usually
based on a joint distribution of both input and output variables
\citep{Rabiner:89} or a conditional distribution of the output given
the input \citep{Lafferty:01}. Therefore this paradigm offers a
flexible probabilistic framework that can naturally facilitate:
hidden variables that capture latent semantics such as a generative
hierarchy \citep{Quattoni:04,Zhu:08c}; Bayesian regularization that
imposes desirable biases such as sparseness
\citep{Lee:06,Wainwright:06,Andrew:07}; and Bayesian prediction
based on combining predictions across all values of model parameters
(i.e., model averaging), which can reduce the risk of overfitting.
On the other hand, the margin-based structured prediction models
leverage the maximum margin principle and convex optimization
formulation underlying the support vector machines, and concentrate
directly on the input-output mapping
\citep{Taskar:03,Altun:03,Tsochantaridis:04}. In principle, this
approach can lead to a robust decision boundary due to the dual
sparseness (i.e., depending on only a few support vectors) and
global optimality of the learned model. However, although arguably a
more desirable paradigm for training highly discriminative
structured prediction models in a number of application contexts,
the lack of a straightforward probabilistic interpretation of the
maximum-margin models makes them unable to offer the same
flexibilities of likelihood-based models discussed above.

For example, for domains with complex feature space, it is often
desirable to pursue a ``sparse" representation of the model that
leaves out irrelevant features. In likelihood-based estimation,
sparse model fitting has been extensively studied. A commonly used
strategy is to add an $L_1$-penalty to the likelihood function,
which can also be viewed as a MAP estimation under a Laplace prior.
However, little progress has been made so far on learning sparse MNs
or log-linear models in general based on the maximum margin
principle. While sparsity has been pursued in maximum margin
learning of certain discriminative models such as SVM that are
``unstructured" (i.e., with a univariate output), by using
$L_1$-regularization \citep{Bennett:92} or by adding a cardinality
constraint \citep{Chan:07}, generalization of these techniques to
structured output space turns out to be extremely non-trivial, as we
discuss later in this paper. There is also very little theoretical
analysis on the performance guarantee of margin-based models under
direct $L_1$-regularization. Our empirical results as shown in this
paper suggest that an $L_1$-regularized estimation, especially the
likelihood based estimation, 
can be unrobust. Discarding the features that are not completely
irrelevant can potentially hurt generalization ability.

In this paper, we propose a general theory of maximum entropy
discrimination Markov networks (MaxEnDNet, or simply MEDN) for
structured input/output learning and prediction. This formalism
offers a formal paradigm for integrating both generative and
discriminative principles and the Bayesian regularization techniques
for learning structured prediction models. It integrates the spirit
of maximum margin learning from SVM, the design of discriminative
structured prediction model in maximum margin Markov networks
(M$^3$N), and the ideas of entropy regularization and model
averaging in maximum entropy discrimination
methods~\citep{Jaakkola:99}. It allows one to learn a distribution
of maximum margin structured prediction models that offers a wide
range of important advantages over conventional models such as
M$^3$N, including more robust prediction due to an averaging 
prediction-function based on the learned distribution of models,
Bayesian-style regularization that can lead to a model that is
simultaneous primal and dual sparse, and allowance of hidden
variables and semi-supervised learning based on partially labeled data.

While the formalism of MaxEnDNet is extremely general, our main
focus and contributions of this paper will be concentrated on the
following results. We will formally define the MaxEnDNet as solving
a generalized entropy optimization problem subject to expected
margin constraints due to the training data, and under an arbitrary
prior of feature coefficients; and we offer a general close-form
solution to this problem. An interesting insight immediately follows
this general solution is that, a trivial assumption on the prior
distribution of the coefficients, i.e., a standard normal, reduces
the linear MaxEnDNet to the standard M$^3$N, as shown in
Theorem~\ref{thrm_gaussianmedn}. This understanding opens the way to
use different priors for MaxEnDNet to achieve more interesting
regularization effects. We show that, by using a Laplace prior for
the feature coefficients, the resulting LapMEDN is effectively an
M$^3$N that is not only dual sparse (i.e., defined by a few support
vectors), but also primal sparse (i.e., shrinkage on coefficients
corresponding to irrelevant features). We develop a novel
variational approximate learning method for the LapMEDN, which
leverages on the hierarchical representation of the Laplace prior
\citep{Figueiredo:03} and the reducibility of MaxEnDNet to M$^3$N,
and combines the variation Bayesian technique with existing convex
optimization algorithms developed for M$^3$N
\citep{Taskar:03,Bartlett:04,Ratliff:07}. We also provide a formal
analysis of the generalization error of the MaxEnDNet, and prove a
novel PAC-Bayes bound on the structured prediction error by
MaxEnDNet. We performed a thorough comparison of the Laplace
MaxEnDNet with a competing methods, including M$^3$N (i.e., the
Gaussian MaxEnDNet), $L_1$-regularized M$^3$N \footnote{This model
has not yet been reported in the literature, and represents another
new extension of the M$^3$N, which we will present in a separate
paper in detail.}, CRFs, $L_1$-regularized CRFs, and
$L_2$-regularized CRFs, on both synthetic and real structured
input/output data. The Laplace MaxEnDNet exhibits mostly superior,
and sometimes comparable performance in all scenarios been tested.

The rest of the paper is structured as follows. In the next section,
we review the basic structured prediction formalism and set the
stage for our model. Section 3 presents the general theory of
maximum entropy discrimination Markov networks and some basic
theoretical results, followed by two instantiations of the general
MaxEnDNet, the Gaussian MaxEnDNet and the Laplace MaxEnDNet. Section
4 offers a detailed discussion of the primal and dual sparsity
property of Laplace MaxEnDNet. Section 5 presents a novel iterative
learning algorithm based on variational approximation and convex
optimization. In Section 6, we briefly discuss the generalization
bound of MaxEnDNet. Then, we show empirical results on both
synthetic and real OCR and web data extraction data sets in Section
7. Section 8 discusses some related work and Section 9 concludes
this paper.

\section{Preliminaries}

In structured prediction problems such as natural language parsing,
image annotation, or DNA decoding, one aims to learn a function
$h:\mathcal{X} \to \mathcal{Y}$ that maps a structured input $\xv
\in \mathcal{X}$, e.g., a sentence or an image, to a structured
output $\yv \in \mathcal{Y}$, e.g., a sentence parsing or a scene
annotation, where, unlike a standard classification problem, $\yv$
is a multivariate prediction consisting of multiple labeling
elements. Let $L$ denote the cardinality of the output, and $m_l$
where $l=1, \ldots, L$ denote the arity of each element, then
$\mathcal{Y}= \mathcal{Y}_1 \times \cdots \times \mathcal{Y}_L$ with
$\mathcal{Y}_l = \{a_1, \ldots, a_{m_l}\}$ represents a
combinatorial space of structured interpretations of the multi-facet
objects in the inputs. For example, $\mathcal{Y}$ could correspond
to the space of all possible instantiations of the parse trees of a
sentence, or the space of all possible ways of labeling entities
over some segmentation of an image. The prediction $\yv \equiv (y_1,
\ldots, y_L)$ is {\it structured} because each individual label $y_l
\in \mathcal{Y}_l$ within $\yv$ must be determined in the context of
other labels $y_{l' \neq l}$, rather than independently as in
classification, in order to arrive at a globally satisfactory and
consistent prediction.

Let $F: \mathcal{X} \times \mathcal{Y} \to \mathbb{R}$ represent a
discriminant function over the input-output pairs from which one can
define the predictive function, and let $\mathcal{H}$ denote the
space of all possible $F$. A common choice of $F$ is a linear model,
$F(\xv, \yv; \wv) = g(\wv ^\top \fv(\xv,\yv))$, where $\fv=[f_1
\ldots f_K]^\top$ is a $K$-dimensional column vector of the feature
functions $f_k: \mathcal{X} \times \mathcal{Y} \to \mathbb{R}$, and
$\wv=[w_1 \ldots w_K]^\top$ is the corresponding vector of the
weights of the feature functions. Typically, a structured prediction
model chooses an optimal estimate $\wv^\star$ by minimizing some
loss function $J(\wv)$, and defines a predictive function in terms
of an optimization problem that maximizes $F(\ \cdot \ ; \wv^\star)$
over the response variable $\yv$ given an input $\xv$:
\begin{equation}
h_0(\xv; \wv^\star) = \arg \max_{\yv \in \mathcal{Y}(\xv)} F(\xv,
\yv; \wv^\star), \label{SP}
\end{equation}
where $\mathcal{Y}(\xv) \subseteq \mathcal{Y}$ is the feasible
subset of structured labels for the input $\xv$. Here, we assume
that $\mathcal{Y}(\xv)$ is finite for any $\xv$.

Depending on the specific choice of $F(\ \cdot \ ; \wv)$ (e.g.,
linear, or log linear), and of the loss function $J(\wv)$ for
estimating the parameter $\wv^\star$ (e.g., likelihood, or margin),
incarnations of the general structured prediction formalism
described above can be seen in classical generative models such as
the HMM~\citep{Rabiner:89} where $g(\cdot)$ can be an exponential
family distribution function and $J(\wv)$ is the joint likelihood of
the input and its labeling; and in recent discriminative models such
as the CRFs \citep{Lafferty:01}, where $g(\cdot)$ is a Boltzmann
machine and $J(\wv)$ is the conditional likelihood of the structured
labeling given input; and the M$^3$N~\citep{Taskar:03}, where
$g(\cdot)$ is an identity function and $J(\wv)$ is the margin
between the true labeling and any other feasible labeling in
$\mathcal{Y}(\xv)$. Our approach toward a more general
discriminative training is based on a maximum entropy principle that
allows an elegant combination of the discriminative maximum margin
learning with the generative Bayesian regularization and
hierarchical modeling, and we consider the more general problem of
finding a distribution over $\mathcal{H}$ that enables a convex
combination of discriminant functions for robust structured
prediction.

Before delving into the exposition of the proposed approach, we end
this section with a brief recapitulation of the basic M$^3$N, upon
which the proposed approach is built. Under a max-margin
framework, given a set of fully observed training data $\mathcal{D}
= \lbrace \langle \xv^i, \yv^i \rangle \rbrace_{i=1}^N$, we obtain a
point estimate of the weight vector $\wv$ by solving the
following max-margin problem P0~\citep{Taskar:03}:\\[-0.4cm]
\begin{eqnarray}
{\rm P0 ~(M^3N):} & & \min_{\wv, \xiv}~\frac{1}{2} \Vert
\wv \Vert^2 + C \sum_{i=1}^N \xi_i  \nonumber \\
\mathrm{s.t.} ~ \forall i, \forall \yv \neq \yv^i: & & \wv^\top
\Delta \fv_i(\yv) \geq \Delta \ell_i(\yv) - \xi_i, ~  \xi_i \ge 0
~, \nonumber
\end{eqnarray}
where $\Delta \fv_i(\yv) = \fv(\xv^i, \yv^i) - \fv(\xv^i, \yv)$ and
$\Delta F_i(\yv; \wv) = \wv^\top \Delta \fv_i(\yv)$ is the ``margin"
between the true label $\yv^i$ and a prediction $\yv$, $\Delta
\ell_i(\yv)$ is a loss function with respect to $\yv^i$, and $\xi_i$
represents a slack variable that absorbs errors in the training
data. Various loss functions have been proposed in the
literature~\citep{Tsochantaridis:04}. In this paper, we adopt the
\textit{hamming loss} used in \citep{Taskar:03}: $\Delta \ell_i(\yv)
= \sum_{j=1}^{L} \indicator(y_j \neq y_j^i)$, where
$\indicator(\cdot)$ is an indicator function that equals to one if
the argument is true and zero otherwise. The optimization problem P0
is intractable because the feasible space for $\wv$,
\[\mathcal{F}_0=\big\{\wv  : \wv^\top \Delta \fv_i(\yv) \geq \Delta
\ell_i(\yv) - \xi_i; \  \forall i, \forall \yv \neq \yv^i \big\},\]
is defined by $O(N|\mathcal{Y}|)$ number of constraints, and
$\mathcal{Y}$ itself is exponential to the size of the input $\xv$.
Exploring sparse dependencies among individual labels $y_l$ in
$\yv$, as reflected in the specific design of the feature functions
(e.g., based on pair-wise labeling potentials in a pair-wise Markov
network), and the convex duality of the objective, efficient
optimization algorithms based on
cutting-plane~\citep{Tsochantaridis:04} or
message-passing~\citep{Taskar:03} have been proposed to obtain an
approximate optimum solution to P0. As described shortly, these
algorithms can be directly employed as subroutines in solving our
proposed model.

\section{Maximum Entropy Discrimination Markov Networks}

Instead of learning a point estimator of $\wv$ as in M$^3$N, in this
paper, we take a Bayesian-style approach and learn a distribution
$p(\wv)$, in a max-margin manner. For prediction, we employ a convex
combination of all possible models $F(\ \cdot \ ; \wv) \in \cal H$
based on $p(\wv)$, that is:
\begin{equation}
h_1(\xv) = \mathrm{arg}\max_{\yv \in \mathcal{Y}(\xv)} \int
p(\wv) F(\xv, \yv; \wv) \, \ud \wv \,. \label{BM3N}
\end{equation}

Now, the open question underlying this averaging prediction rule is
how we can devise an appropriate loss function and constraints over
$p(\wv)$, in a similar spirit as the margin-based scheme over $\wv$
in P0, that lead to an optimum estimate of $p(\wv)$. In the sequel,
we present {\it Maximum Entropy Discrimination Markov Networks}
(MaxEnDNet, or MEDN), a novel framework that facilitates the
estimation of a Bayesian-style regularized {\it distribution} of
M$^3$Ns defined by $p(\wv)$. As we show below, this new
Bayesian-style max-margin learning formalism offers several
advantages such as simultaneous primal and dual sparsity,
PAC-Bayesian generalization guarantee, and estimation robustness.
Note that the MaxEnDNet is different from the traditional Bayesian
methods for discriminative structured prediction such as the
Bayesian CRFs \citep{Qi:05}, where the likelihood function is well
defined. Here, our approach is of a ``Bayesian-style" because it
learns and uses a ``posterior" distribution of all predictive models
instead of choosing one model according to some criterion, but the
learning algorithm is not based on the Bayes theorem, but a maximum
entropy principle that biases towards a posterior that makes less
additional assumptions over a given prior over the predictive
models.

\subsection{Structured Maximum Entropy Discrimination}

Given a training set $\mathcal{D}$ of structured input-output pairs,
analogous to the feasible space $\mathcal{F}_0$ for the weight
vector $\wv$ in a standard M$^3$N (c.f., problem P0), we define the
feasible subspace $\mathcal{F}_1$ for the weight distribution
$p(\wv)$ by a set of {\it expected} margin constraints:

\begin{displaymath}
\mathcal{F}_1 = \Big \lbrace p(\wv):\int p(\wv) \lbrack \Delta
F_i(\yv; \wv) - \Delta \ell_i(\yv) \rbrack \, \ud \wv  \geq
-\xi_i,~\forall i, \forall \yv \neq \yv^i \Big \rbrace.
\end{displaymath}


We learn the optimum $p(\wv)$ from $\mathcal{F}_1$ based on a {\it
structured maximum entropy discrimination principle} generalized
from~\citep{Jaakkola:99}. Under this principle, the optimum $p(\wv)$
corresponds to the distribution that minimizes its relative entropy
with respect to some chosen prior $p_0$, as measured by the
Kullback-Leibler divergence between $p$ and $p_0$: $KL(p||p_0) =
\langle \log (p/p_0) \rangle_p$, where $\langle \cdot \rangle_p$
denotes the expectations with respect to $p$. If $p_0$ is uniform,
then minimizing this KL-divergence is equivalent to maximizing the
entropy $H(p)=- \langle \log p \rangle_p$. A natural information
theoretic interpretation of this formulation is that we favor a
distribution over the hypothesis class $\cal H$ that bears minimum
assumptions among all feasible distributions in $\mathcal{F}_1$. The
$p_0$ is a regularizer that introduces an appropriate bias, if
necessary.

To accommodate non-separable cases in the discriminative prediction
problem, instead of minimizing the usual KL, we optimize
the {\it generalized entropy} \citep{Dudik:07,Lebanon:01}, or a
regularized KL-divergence, $KL(p(\wv) || p_0(\wv) ) + U(\xiv)$,
where $U(\xiv)$ is a closed proper convex function over the slack
variables. This term can be understood as an additional
``potential'' in the maximum entropy principle. Putting everything
together, we can now state a general formalism based on the
following Maximum Entropy Discrimination Markov Network framework:

\begin{definition}{\bf (Maximum Entropy Discrimination Markov Networks)}
Given training data $\mathcal{D}=\lbrace \langle \xv^i, \yv^i
\rangle \rbrace_{i=1}^N$, a chosen form of discriminant
function~$F(\xv, \yv; \wv)$, a loss function $\Delta \ell(\yv)$, and
an ensuing feasible subspace $\mathcal{F}_1$ (defined above) for
parameter distribution $p(\wv)$, the MaxEnDNet model that leads to a
prediction function of the form of Eq.
(\ref{BM3N}) is defined by the following generalized relative entropy minimization with respect to a parameter prior $p_0(\wv)$:\\[-0.6cm]
\begin{eqnarray}
\hspace{-2cm} {\rm P1 ~(MaxEnDNet)}: & & \min_{p(\wv),
\xi}~KL(p(\wv) || p_0(\wv) ) +
U(\xiv)  \nonumber \\
\hspace{-2cm}& & ~ \mathrm{s.t.} ~ ~ p(\wv)  \in \mathcal{F}_1, ~
\xi_i \ge 0, \forall i. \nonumber
\end{eqnarray}
\end{definition}

The P1 defined above is a variational optimization problem over
$p(\wv)$ in a subspace of valid parameter distributions. Since both
the KL and the function $U$ in P1 are convex, and the constraints in
$\mathcal{F}_1$ are linear, P1 is a convex program. In addition, the
expectations $\langle F(\xv, \yv; \wv) \rangle_{p(\wv)}$ are
required to be bounded in order for $F$ to be a meaningful model.
Thus, the problem P1 satisfies the {\it Slater's
condition}\footnote{Since $\langle F(\xv, \yv; \wv)
\rangle_{p(\wv)}$ are bounded and $\xi_i \geq 0$, there always
exists a $\xi$, which is large enough to make the pair $(p(\wv),
\xi)$ satisfy the Slater's condition.}~\citep[chap.~5]{Boyd:04},
which together with the convexity make P1 enjoy nice properties,
such as strong duality and the existence of solutions. The problem
P1 can be solved via applying the calculus of variations to the
Lagrangian to obtain a variational extremum, followed by a dual
transformation of P1. We state the main results below as a theorem,
followed by a brief proof that lends many insights into the solution
to P1 which we will explore in subsequent analysis.

\begin{theorem}[Solution to MaxEnDNet]\label{thrm_medn}
The variational optimization problem P1 underlying the MaxEnDNet
gives rise to the following optimum distribution of Markov network
parameters $\wv$:
\begin{eqnarray}\label{wdist}
p(\wv) = \frac{1}{Z(\alphav)} p_0(\wv) \exp \Big \lbrace \sum_{i,
\yv \neq \yv^i} \alpha_i(\yv) \lbrack \Delta F_{i}(\yv; \wv) -
\Delta \ell_{i}(\yv) \rbrack \Big \rbrace ,
\end{eqnarray}

\noindent where $Z(\alphav)$ is a normalization factor and the
Lagrangian multipliers $\alpha_i(\yv)$ (corresponding to the
constraints in $\mathcal{F}_1$) can be obtained by solving the dual
problem of P1:
\begin{eqnarray}
{\rm D1:} &&\max_{\alphav}~ -\log Z(\alphav) - U^\star (\alphav) \nonumber \\
&&\mathrm{s.t.}~~\alpha_i(\yv) \geq 0, ~\forall i,~ \forall \yv \neq
\yv^i \nonumber
\end{eqnarray}
where $U^\star(\cdot)$ is the conjugate of the slack function
$U(\cdot)$, i.e.,  $U^\star(\alpha)=\sup_{\xi} \big( \sum_{i, \yv
\neq \yv^i} \alpha_i(\yv) \xi_i - U(\xi) \big)$.
\end{theorem}

\begin{proof}
({\it sketch}) Since the problem P1 is a convex program and
satisfies the Slater's condition, we can form a Lagrange function,
whose saddle point gives the optimal solution of P1 and D1, by
introducing a non-negative dual variable $\alpha_i(\yv)$ for each
constraint in $\mathcal{F}_1$ and another non-negative dual variable
$c$ for the normalization
constraint $\int p(\wv) \, \ud \wv \, = 1$. 
Details are deferred to Appendix B.1.
\end{proof}

Since the problem P1 is a convex program and satisfies the Slater's
condition, the saddle point of the Lagrange function is the KKT
point of P1. From the KKT conditions~\citep[chap.~5]{Boyd:04}, it
can be shown that the above solution enjoys {\it dual sparsity},
that is, only a few Lagrangian multipliers will be non-zero, which
correspond to the active constraints whose equality holds, analogous
to the support vectors in SVM. Thus MaxEnDNet enjoys a similar
generalization property as the M$^3$N and SVM due to the the small
``effective size" of the margin constraints. But it is important to
realize that this does not mean that the learned model is
``primal-sparse", i.e., only a few elements in the weight vector
$\wv$ are non-zero. We will return to this point in
Section~\ref{sec:shrinkageanalysis}.

For a closed proper convex function $\phi(\mu)$, its conjugate is
defined as $\phi^\star (\nu) = \sup_{\mu} \lbrack \nu^\top \mu -
\phi(\mu) \rbrack$. In the problem D1, by convex
duality~\citep{Boyd:04}, the log normalizer $\log Z(\alpha)$
can be shown to be the conjugate of the KL-divergence. If the
slack function is $U(\xi) = C\Vert \xi \Vert = C\sum_i \xi_i$, it
is easy to show that $U^\star(\alpha) = \indicator_{\D
\infty}(\sum_{\yv} \alpha_i(\yv) \leq C,~\forall i)$, where
$\indicator_{\D \infty}(\cdot)$ is a function that equals to zero
when its argument holds true and infinity otherwise. Here, the
inequality corresponds to the trivial solution $\xi=0$, that is,
the training data are perfectly separative. Ignoring this
inequality does not affect the solution since the special case
$\xi=0$ is still included. Thus, the Lagrangian multipliers
$\alpha_i(\yv)$ in the dual problem D1 comply with the set of
constraints that $\sum_{\yv} \alpha_i(\yv) = C,~\forall i$.
Another example is $U(\xi)=KL(p(\xi) || p_0(\xi))$ by introducing
uncertainty on the slack variables \citep{Jaakkola:99}. In this
case, expectations with respect to $p(\xi)$ are taken on both
sides of all the constraints in $\mathcal{F}_1$. Take the duality,
and the dual function of $U$ is another log normalizer. More
details can be found in \citep{Jaakkola:99}. Some other $U$
functions and their dual functions are studied in
\citep{Lebanon:01,Dudik:07}.

Unlike most extant structured discriminative models including the
highly successful M$^3$N, which rely on a point estimator of the
parameters, the MaxEnDNet model derived above gives an optimum
parameter distribution, which is used to make prediction via the
rule (\ref{BM3N}). Indeed, as we will show shortly, the MaxEnDNet is
strictly more general than the M$^3$N and subsumes the later as a
special case. But more importantly, the MaxEnDNet in its full
generality offers a number of important advantages while retaining
all the merits of the M$^3$N. {\bf First}, MaxEnDNet admits a prior
that can be designed to introduce useful regularization effects,
such as a primal sparsity bias. {\bf Second}, the MaxEnDNet
prediction is based on model averaging and therefore enjoys a
desirable smoothing effect, with a uniform
convergence bound on generalization error. 
{\bf Third}, MaxEnDNet offers a principled way to incorporate {\it
hidden} generative models underlying the structured predictions, but
allows the predictive model to be discriminatively trained based on
partially labeled data. In the sequel, we analyze the first two
points in detail; exploration of the third point is beyond the scope
of this paper, and can be found in~\citep{Zhu:08b}, where a {\it
partially observed} MaxEnDNet (PoMEN) is developed, which combines
(possibly latent) generative model and discriminative training for
structured prediction.

\subsection{Gaussian MaxEnDNet}

As Eq. (\ref{wdist}) suggests, different choices of the parameter
prior can lead to different MaxEnDNet models for predictive
parameter distribution. In this subsection and the following one, we
explore a few common choices, e.g., Gaussian and Laplace priors.

We first show that, when the parameter prior is set to be a standard
normal, MaxEnDNet leads to a predictor that is identical to that of
the M$^3$N. This somewhat surprising reduction offers an important
insight for understanding the property of MaxEnDNet. Indeed this
result should not be totally unexpected given the striking
isomorphisms of the opt-problem P1, the feasible space
$\mathcal{F}_1$, and the predictive function $h_1$ underlying a
MaxEnDNet, to their counterparts P0, $\mathcal{F}_0$, and $h_0$,
respectively, underlying an M$^3$N. The following theorem makes our
claim explicit.

\begin{theorem}[Gaussian MaxEnDNet: Reduction of MEDN to
M$^3$N]\label{thrm_gaussianmedn} Assuming \\
$F(\xv, \yv; \wv) = \wv^\top \fv(\xv, \yv)$, $U(\xi)= C \sum_i
\xi_{i}$, \textit{and}~$p_0(\wv) = \mathcal{N}(\wv |0, I)$, where
$I$ denotes an identity matrix, then the posterior distribution is
$p(\wv) = \mathcal{N}(\wv|\mu, I)$, where $\mu = \sum_{i,\yv \neq
\yv^i} \alpha_i(\yv) \Delta \fv_{i}(\yv)$, and the Lagrangian
multipliers $\alpha_i(\yv)$ in $p(\wv)$ are obtained by solving the
following dual problem, which is isomorphic to the dual form of the
M$^3$N:
\begin{eqnarray}
&&\max_{\alpha}~ \sum_{i, \yv \neq \yv^i} \alpha_i(\yv) \Delta
\ell_{i}(\yv) - \frac{1}{2} \Vert \sum_{i, \yv \neq \yv^i}
\alpha_i(\yv) \Delta \fv_{i}(\yv) \Vert^2 \nonumber \\
&&\mathrm{s.t.}~~\sum_{\yv \neq \yv^i} \alpha_i(\yv) =
C;~\alpha_i(\yv) \geq 0, ~\forall i,~ \forall \yv \neq \yv^i,
\nonumber
\end{eqnarray}

\noindent where $\Delta \fv_{i}(\yv) = \fv(\xv^i, \yv^i) -
\fv(\xv^i, \yv)$ as in P0. When applied to $h_1$, $p(\wv)$ leads to
a predictive function that is identical to $h_0(\xv; \wv)$ given by
Eq. (\ref{SP}).
\end{theorem}

\begin{proof}
See Appendix B.2 for details.
\end{proof}

The above theorem is stated in the duality form. We can also show
the following equivalence in the primal form.

\begin{corollary}\label{cor_primal_gaussianmedn}
Under the same assumptions as in Theorem~\ref{thrm_gaussianmedn},
the mean $\mu$ of the posterior distribution $p(\wv)$ under a
Gaussian MaxEnDNet is obtained by solving the following primal
problem:
\begin{eqnarray}
&&\min_{\mu,\xi}~\frac{1}{2} \mu^\top \mu + C \sum_{i=1}^N \xi_i \nonumber \\
&&\mathrm{s.t.}~~\mu^\top \Delta \fv_i(\yv) \geq \Delta
\ell_{i}(\yv) - \xi_i;~\xi_i \geq 0,~~\forall i,~\forall \yv \neq
\yv^i. \nonumber
\end{eqnarray}
\end{corollary}

\begin{proof}
See Appendix B.3 for details.
\end{proof}

Theorem~\ref{thrm_gaussianmedn} and
Corollary~\ref{cor_primal_gaussianmedn} both show that in the
supervised learning setting, the M$^3$N is a special case of
MaxEnDNet when the slack function is linear and the parameter prior
is a standard normal. As we shall see later, this connection renders
many existing techniques for solving the M$^3$N directly applicable
for solving the MaxEnDNet.

\subsection{Laplace MaxEnDNet}\label{sec:lapmedn}

Recent trends in pursuing ``sparse" graphical models has led to the
emergence of regularized version of CRFs~\citep{Andrew:07} and
Markov networks~\citep{Lee:06,Wainwright:06}. Interestingly, while
such extensions have been successfully implemented by several
authors in maximum likelihood learning of various sparse graphical
models, they have not yet been explored in the context of maximum
margin learning. Such a gap is not merely due to a negligence.
Indeed, learning a sparse M$^3$N can be significantly harder as we
discuss below.

One possible way to learn a sparse M$^3$N is to adopt the strategy
of $L_1$-SVM~\citep{Bennett:92,Zhu:04} and directly use an $L_1$
instead of the $L_2$-norm of $\wv$ in the loss function (see
appendix A for a detailed description of this formulation and the
duality derivation). However, the primal problem of an
$L_1$-regularized M$^3$N is not directly solvable by re-formulating
it as an LP problem due to the exponential number of constraints;
solving the dual problem, which now has only a polynomial number of
constraints as in the dual of M$^3$N, is still non-trivial due to
the complicated form of the constraints as shown in appendix A. The
constraint generation methods are possible. However, although such
methods \citep{Tsochantaridis:04} have been shown to be efficient
for solving the QP problem in the standard M$^3$N, our preliminary
empirical results show that such a scheme with an LP solver for the
$L_1$-regularized M$^3$N can be extremely expensive for a
non-trivial real data set. Another possible solution is the gradient
descent methods \citep{Ratliff:07} with a projection to $L_1$-ball
\citep{Duchi:08}.

The MaxEnDNet interpretation of the M$^3$N offers an alternative
strategy that resembles Bayesian
regularization~\citep{Tipping:01,Kaban:07} in maximum likelihood
estimation, where shrinkage effects can be introduced by appropriate
priors over the model parameters. As Theorem~\ref{thrm_gaussianmedn}
reveals, an M$^3$N corresponds to a Gaussian MaxEnDNet that admits a
standard normal prior for the weight vector $\wv$. According to the
standard Bayesian regularization theory, to achieve a sparse
estimate of a model, in the posterior distribution of the feature
weights, the weights of irrelevant features should peak around zero
with very small variances. However, the isotropy of the variances in
all dimensions of the feature space under a standard normal prior
makes it infeasible for the resulting M$^3$N to adjust the variances
in different dimensions to fit a sparse model. Alternatively, now we
employ a Laplace prior for $\wv$ to learn a Laplace MaxEnDNet. We
show in the sequel that, the parameter posterior $p(\wv)$ under a
Laplace MaxEnDNet has a shrinkage effect on small weights, which is
similar to directly applying an $L_1$-regularizer on an M$^3$N.
Although exact learning of a Laplace MaxEnDNet is also intractable,
we show that this model can be efficiently approximated by a
variational inference procedure based on existing methods.

The Laplace prior of $\wv$ is expressed as $p_0(\wv) = \prod_{k=1}^K
\frac{\sqrt{\lambda}}{2}e^{-\sqrt{\lambda} |w_k|} = \big(
\frac{\sqrt{\lambda}}{2} \big)^K e^{-\sqrt{\lambda} \Vert \wv
\Vert}$. This density function is heavy tailed and peaked at zero;
thus, it encodes a prior belief that the distribution of $\wv$ is
strongly peaked around zero. Another nice property of the Laplace
density is that it is log-concave, or the negative logarithm is
convex, which can be exploited to obtain a convex estimation problem
analogous to LASSO \citep{Tibshirani:96}.

\begin{theorem}[Laplace MaxEnDNet: a sparse
M$^3$N]\label{thrm_lapmaxendnet} Assuming $F(\xv, \yv; \wv) =
\wv^\top \fv(\xv, \yv)$, $U(\xi)= C \sum_i \xi_{i}$,
\textit{and}~$p_0(\wv) = \prod_{k=1}^K
\frac{\sqrt{\lambda}}{2}e^{-\sqrt{\lambda} |w_k|} = \big(
\frac{\sqrt{\lambda}}{2} \big)^K e^{-\sqrt{\lambda} \Vert \wv
\Vert}$, then the Lagrangian multipliers $\alpha_i(\yv)$ in $p(\wv)$
(as defined in Theorem 2) are obtained by solving the following dual
problem:
\begin{eqnarray}
&& \max_{\alpha}~ \sum_{i,\yv \neq \yv^i} \alpha_i(\yv) \Delta
\ell_{i}(\yv) - \sum_{k=1}^K \log \frac{\lambda}{\lambda - \eta_k^2}
\nonumber \\
&&\mathrm{s.t.}~~\sum_{\yv \neq \yv^i} \alpha_i(\yv) =
C;~\alpha_i(\yv) \geq 0, ~\forall i,~ \forall \yv \neq \yv^i.
\nonumber
\end{eqnarray}
\noindent where $\eta_k = \sum_{i,\yv \neq \yv^i} \alpha_i(\yv)
\Delta \fv_i^{\D k}(\yv)$, and $\Delta \fv_i^{\D k}(\yv) =
f_k(\xv^i, \yv^i) - f_k(\xv^i, \yv)$ represents the $k$th component
of $\Delta \fv_i(\yv)$. Furthermore, constraints $\eta_k^2 <
\lambda,~\forall k$, must be satisfied.
\end{theorem}

Since several intermediate results from the proof of this Theorem
will be used in subsequent presentations, we provide the complete
proof below. Our proof is based on a hierarchical representation of
the Laplace prior. As noted in \citep{Figueiredo:03}, the Laplace
distribution $p(w) = \frac{\sqrt{\lambda}}{2}
e^{-\sqrt{\lambda}|w|}$ is equivalent to a two-layer hierarchical
Gaussian-exponential model, where $w$ follows a zero-mean Gaussian
distribution $p(w| \tau) = \mathcal{N}(w|0, \tau)$ and the variance
$\tau$ admits an exponential hyper-prior density,
\begin{displaymath}
p(\tau | \lambda) = \frac{\lambda}{2} \exp \big \lbrace
-\frac{\lambda}{2} \tau \big \rbrace,~~\mathrm{for}~\tau \geq 0.
\end{displaymath}
\noindent This alternative form straightforwardly leads to the
following new representation of our multivariate Laplace prior for
the parameter vector $\wv$ in MaxEnDNet:
\begin{equation}
p_0(\wv) = \prod_{k=1}^K p_0(w_k) = \prod_{k=1}^K \int p(w_k |
\tau_k) p(\tau_k | \lambda) \, \ud \tau_k \, = \int p(\wv | \tauv)
p(\tauv | \lambda) \, \ud \tauv, \label{hLap}
\end{equation}
\noindent where $p(\wv | \tauv) = \prod_{k=1}^K p(w_k | \tau_k)$ and
$p(\tauv | \lambda) = \prod_{k=1}^K p(\tau_k | \lambda)$ represent
multivariate Gaussian and exponential, respectively, and $\ud \tauv
\triangleq \ud \tau_1 \cdots \ud \tau_K$.

\begin{proof} ({\it of Theorem \ref{thrm_lapmaxendnet}})
Substitute the hierarchical representation of the Laplace prior (Eq.
\ref{hLap}) into $p(\wv)$ in Theorem~\ref{thrm_medn}, and we get the
normalization factor $Z(\alpha)$ as follows,
\setlength\arraycolsep{1pt}{\begin{eqnarray}\label{normfactor_lapmedn}
Z(\alpha) &&= \int \int p(\wv | \tau) p(\tau | \lambda) \, \ud \tau
\, \cdot \exp \lbrace \wv^\top \eta - \sum_{i, \yv \neq \yv^i}
\alpha_i(\yv)
\Delta \ell_{i}(\yv) \rbrace \, \ud \wv \, \nonumber \\
&& = \int p(\tau | \lambda) \int p(\wv | \tau) \cdot \exp \lbrace
\wv^\top \eta - \sum_{i,\yv \neq \yv^i} \alpha_i(\yv)
\Delta \ell_{i}(\yv) \rbrace \, \ud \wv \, \, \ud \tau \, \nonumber \\
&& = \int p(\tau | \lambda) \int \mathcal{N}(\wv | 0, A) \exp
\lbrace \wv^\top \eta - \sum_{i,\yv \neq \yv^i} \alpha_i(\yv)
\Delta \ell_{i}(\yv) \rbrace \, \ud \wv \, \, \ud \tau \, \nonumber \\
&& = \int p(\tau | \lambda) \exp \lbrace \frac{1}{2} \eta^\top A
\eta - \sum_{i,\yv \neq \yv^i} \alpha_i(\yv)
\Delta \ell_{i}(\yv) \rbrace \ud \tau \, \nonumber \\
&& = \exp \lbrace -\sum_{i,\yv \neq \yv^i} \alpha_i(\yv)
\Delta \ell_{i}(\yv) \rbrace \prod_{k=1}^K \int \frac{\lambda}{2} \exp ( - \frac{\lambda}{2} \tau_k ) \exp ( \frac{1}{2} \eta_k^2 \tau_k ) \ud \tau_k \, \nonumber \\
&& = \exp \lbrace -\sum_{i,\yv \neq \yv^i} \alpha_i(\yv) \Delta
\ell_{i}(\yv) \rbrace \prod_{k=1}^K \frac{\lambda}{\lambda
-\eta_k^2}, \label{e0}
\end{eqnarray}}

\noindent where $A = {\rm diag}(\tau_k)$ is a diagonal matrix and
$\eta$ is a column vector with $\eta_k$ defined as in
Theorem~\ref{thrm_lapmaxendnet}. The last equality is due to the
moment generating function of an exponential distribution. The
constraint $\eta_k^2 < \lambda,~\forall k$ is needed in this
derivation to avoid the integration going infinity. Substituting the
normalization factor derived above into the general dual problem D1
in Theorem~\ref{thrm_medn}, and using the same argument of the
convex conjugate of $U(\xi) = C \sum_{i} \xi_i$ as in
Theorem~\ref{thrm_gaussianmedn}, we arrive at the dual problem in
Theorem~\ref{thrm_lapmaxendnet}.
\end{proof}

It can be shown that the dual objective function of Laplace
MaxEnDNet in Theorem~\ref{thrm_lapmaxendnet} is
concave\footnote{$\eta_k^2$ is convex over $\alpha$ because it is
the composition of $f(x) = x^2$ with an affine mapping. So, $\lambda
- \eta_k^2$ is concave and $\log (\lambda - \eta_k^2)$ is also
concave due to the composition rule~\citep{Boyd:04}.}. But since
each $\eta_k$ depends on all the dual variables $\alpha$ and
$\eta_k^2$ appears within a logarithm, the optimization problem
underlying Laplace MaxEnDNet would be very difficult to solve. The
SMO \citep{Taskar:03} and the exponentiated gradient methods
\citep{Bartlett:04} developed for the QP dual problem of M$^3$N
cannot be easily applied here. Thus, we will turn to a variational
approximation method, as shown in Section~\ref{sec:vbmethod}. For
completeness, we end this section with a corollary similar to the
Corollary~\ref{cor_primal_gaussianmedn}, which states the primal
optimization problem underlying the MaxEnDNet with a Laplace prior.
As we shall see, the primal optimization problem in this case is
complicated and provides another perspective of the hardness of
solving the Laplace MaxEnDNet.

\begin{corollary}\label{cor_primal_lapmaxendnet}
Under the same assumptions as in Theorem~\ref{thrm_lapmaxendnet},
the mean~$\mu$~of the posterior distribution~$p(\wv)$ under a
Laplace MaxEnDNet is obtained by solving the following primal
problem:
\begin{eqnarray}
&&\min_{\mu,\xi}~ \sqrt{\lambda} \sum_{k=1}^K \Big( \sqrt{\mu_k^2 +
\frac{1}{\lambda}} - \frac{1}{ \sqrt{\lambda} }
\log \frac{\sqrt{\lambda \mu_k^2+1} + 1}{2}\Big) + C \sum_{i=1}^N \xi_i \nonumber \\
&&\mathrm{s.t.}~~\mu^\top \Delta \fv_i(\yv) \geq \Delta
\ell_{i}(\yv) - \xi_i;~\xi_i \geq 0,~~\forall i,~\forall \yv \neq
\yv^i. \nonumber
\end{eqnarray}
\end{corollary}

\begin{proof}
The proof requires the result of
Corollary~\ref{remark_postshrinkage}. We defer it to Appendix B.4.
\end{proof}

Since the ``norm"\footnote{This is not exactly a norm because the
positive scalability does not hold. But the KL-norm is non-negative
due to the non-negativity of KL-divergence. In fact, by using the
inequality $e^x \geq 1 + x$, we can show that each component $(
\sqrt{\mu_k^2 + \frac{1}{\lambda}} - \frac{1}{ \sqrt{\lambda} } \log
\frac{\sqrt{\lambda \mu_k^2+1} + 1}{2})$ is monotonically increasing
with respect to $\mu_k^2$ and $\Vert \mu \Vert_{KL} \geq K$, where
the equality holds only when $\mu = 0$. Thus, $\Vert \mu \Vert_{KL}$
penalizes large weights. For convenient comparison with the popular
$L_2$ and $L_1$ norms, we call it a KL-norm.}
$$\sum_{k=1}^K \Big( \sqrt{ \mu_k^2 + \frac{1}{\lambda}} -
\frac{1}{\sqrt{\lambda}} \log \frac{\sqrt{\lambda \mu_k^2+1} +
1}{2}\Big) \triangleq \Vert \mu \Vert_{KL}$$ corresponds to the
KL-divergence between $p(\wv)$ and $p_0(\wv)$ under a Laplace
MaxEnDNet, we will refer to it as a {\it KL-norm} and denote it by
$\Vert \cdot \Vert_{KL}$ in the sequel. This KL-norm is different
from the $L_2$-norm as used in M$^3$N, but is closely related to the
$L_1$-norm, which encourages a sparse estimator. In the following
section, we provide a detailed analysis of the sparsity of Laplace
MaxEnDNet resulted from the regularization effect from this norm.

\section{Entropic Regularization and Sparse M$^3$N}
\label{sec:shrinkageanalysis}


Comparing to the structured prediction law $h_0$ due to an M$^3$N,
which enjoys dual sparsity (i.e., few support vectors), the $h_1$
defined by a Laplace MaxEnDNet is not only dual-sparse, but also
primal sparse; that is, features that are insignificant will
experience strong shrinkage on their corresponding weight $w_k$.

The primal sparsity of $h_1$ achieved by the Laplace MaxEnDNet is
due to a shrinkage effect resulting from the {\it Laplacian entropic
regularization}. In this section, we take a close look at this
regularization effect, in comparison with other common regularizers,
such as the $L_2$-norm in M$^3$N (which is equivalent to the
Gaussian MaxEnDNet), and the $L_1$-norm that at least in principle
could be directly applied to M$^3$N. Since our main interest here is
the sparsity of the structured prediction law $h_1$, we examine the
posterior mean under $p(\wv)$ via exact integration. It can be shown
that under a Laplace MaxEnDNet, $p(\wv)$ exhibits the following
posterior shrinkage effect.

\begin{corollary}[Entropic Shrinkage]\label{remark_postshrinkage} The posterior mean of the Laplace
MaxEnDNet has the following form:
\begin{equation}
\langle w_k \rangle_p = \frac{2\eta_k}{\lambda - \eta_k^2},~\forall
1 \leq k \leq K, \label{lap_mean}
\end{equation}
where $\eta_k = \sum_{i,\yv \neq \yv^i} \alpha_i(\yv) (f_k(\xv^i,
\yv^i) - f_k(\xv^i, \yv))$ and $\eta_k^2 < \lambda,~\forall k$.
\end{corollary}

\begin{proof}
Using the integration result in Eq. (\ref{normfactor_lapmedn}), we can get:\\[-0.8cm]

\begin{equation} \frac{\partial{\log
Z}}{\partial \alpha_i(\yv)} = v^\top \Delta
\mathrm{\mathbf{f}}_{i}(\yv) - \Delta \ell_{i}(\yv), \label{e1}
\end{equation}\\[-0.8cm]

\noindent where $v$ is a column vector and $v_k =
\frac{2\eta_k}{\lambda - \eta_k^2},~\forall 1 \leq k \leq K$. An
alternative way to compute the derivatives is using the definition
of $Z:~Z = \int p_0(\wv) \cdot \exp \lbrace \wv^\top \eta -
\sum_{i,\yv \neq \yv^i} \alpha_i(\yv) \Delta \ell_{i}(\yv) \rbrace
\, \ud \wv \,$. We can get:
\begin{equation} \frac{\partial{\log Z}}{\partial \alpha_i(\yv)} =
\langle \wv \rangle_p^\top \Delta \mathrm{\mathbf{f}}_{i}(\yv) -
\Delta \ell_{i}(\yv). \label{e2}
\end{equation}\\[-0.7cm]

\begin{figure}%
\centerline{\includegraphics[width=260pt]{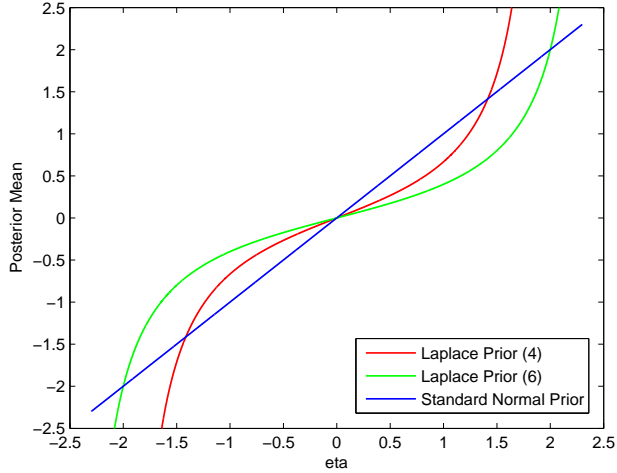}}%
\vspace{-0.5cm} \caption{Posterior means with different priors
against their corresponding $\eta = \sum_{i,\yv \neq \yv^i}
\alpha_i(\yv) \Delta \fv_i(\yv)$. Note that the $\eta$ for different
priors are generally different because of the different dual
parameters. }\label{post_mean} \vspace{-0.5cm}
\end{figure}
Comparing Eqs. (\ref{e1}) and (\ref{e2}), we get $\langle \wv
\rangle_p = v$, that is, $\langle w_k \rangle_p =
\frac{2\eta_k}{\lambda - \eta_k^2},~\forall 1 \leq k \leq K$. The
constraints $\eta_k^2 < \lambda,~\forall k$ are required to get a
finite normalization factor as shown in Eq.
(\ref{normfactor_lapmedn}).
\end{proof}

A plot of the relationship between $\langle w_k \rangle_p$ under a
Laplace MaxEnDNet and the corresponding $\eta_k$
revealed by Corollary \ref{remark_postshrinkage} is shown in
Figure~\ref{post_mean} (for example, the red curve), from which we
can see that, the smaller the $\eta_k$ is, the more shrinkage toward
zero is imposed on $\langle w_k \rangle_p$.

This entropic shrinkage effect on $\wv$ is not present in the
standard M$^3$N, and the Gaussian MaxEnDNet. Recall that by
definition, the vector $\eta \triangleq \sum_{i,\yv} \alpha_i(\yv)
\Delta \fv_i(\yv)$ is determined by the dual parameters
$\alpha_i(\yv)$ obtained by solving a model-specific dual problem.
When the $\alpha_i(\yv)$'s are obtained by solving the dual of the
standard M$^3$N, it can be shown that the optimum point solution of
the parameters $\wv^\star=\eta$. When the $\alpha_i(\yv)$'s are
obtained from the dual of the Gaussian MaxEnDNet, Theorem
\ref{thrm_gaussianmedn} shows that the posterior mean of the
parameters $\langle \wv \rangle_{p_{_{\rm Gaussian}}} = \eta$. (As
we have already pointed out, since these two dual problems are
isomorphic, the $\alpha_i(\yv)$'s for M$^3$N and Gaussian MaxEnDNet
are identical, hence the resulting $\eta$'s are the same.) In both
cases, there is no shrinkage along any particular dimension of the
parameter vector $\wv$ or of the mean vector of $p(\wv)$. Therefore,
although both M$^3$N and Gaussian MaxEnDNet enjoy the dual sparsity,
because the KKT conditions imply that most of the dual parameters
$\alpha_i(\yv)$'s are zero, $\wv^\star$ and $\langle \wv
\rangle_{p_{_{\rm Gaussian}}}$ are not primal sparse. From Eq.
(\ref{lap_mean}), we can conclude that the Laplace MaxEnDNet is also
dual sparse, because its mean $\langle \wv \rangle_{p_{_{\rm
Laplace}}}$ can be uniquely determined by $\eta$. But the shrinkage
effect on different components of the $\langle \wv \rangle_{p_{_{\rm
Laplace}}}$ vector causes $\langle \wv \rangle_{p_{_{\rm Laplace}}}$
to be also primal sparse.

A comparison of the posterior mean estimates of $\wv$ under
MaxEnDNet with three different priors versus their associated $\eta$
is shown in Figure~\ref{post_mean}. The three priors in question
are, a standard normal, a Laplace with $\lambda=4$, and a Laplace
with $\lambda=6$. It can be seen that, under the entropic
regularization with a Laplace prior, the $\langle \wv \rangle_p$
gets shrunk toward zero when $\eta$ is small. The larger the
$\lambda$ value is, the greater the shrinkage effect. For a fixed
$\lambda$, the shape of the shrinkage curve (i.e., the $\langle \wv
\rangle_p - \eta$ curve) is smoothly nonlinear, but no component is
explicitly discarded, that is, no weight is set explicitly to zero.
In contrast, for the Gaussian MaxEnDNet, which is equivalent to the
standard M$^3$N, there is no such a shrinkage effect.

Corollary~\ref{cor_primal_lapmaxendnet} offers another perspective
of how the Laplace MaxEnDNet relates to the $L_1$-norm M$^3$N, which
yields a sparse estimator. Note that as $\lambda$ goes to infinity,
the KL-norm $\Vert \mu \Vert_{KL}$ approaches $\Vert \mu \Vert_1$,
i.e., the $L_1$-norm\footnote{As $\lambda \to \infty$, the logarithm
terms in $\Vert \mu \Vert_{KL}$ disappear because of the fact that
$\frac{\log{x}}{x} \to 0$ when $x \to \infty$.}. This means that the
MaxEnDNet with a Laplace prior will be (nearly) the same as the
$L_1$-M$^3$N if the regularization constant $\lambda$ is large
enough.

\begin{figure*} \centerline{\hfill
\subfigure[]{\includegraphics[width=220pt]{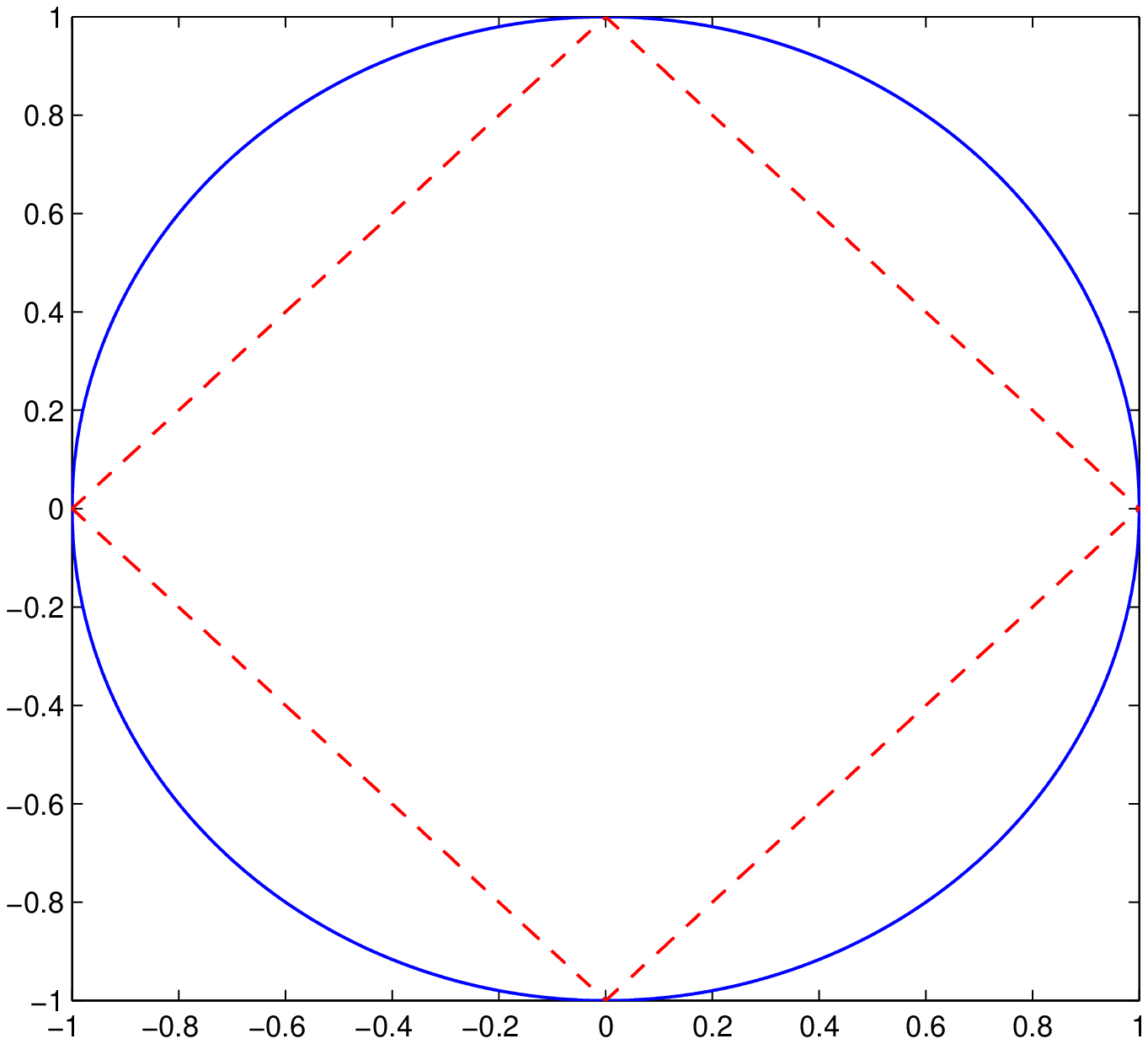}}\hfill
\subfigure[]{\includegraphics[width=220pt]{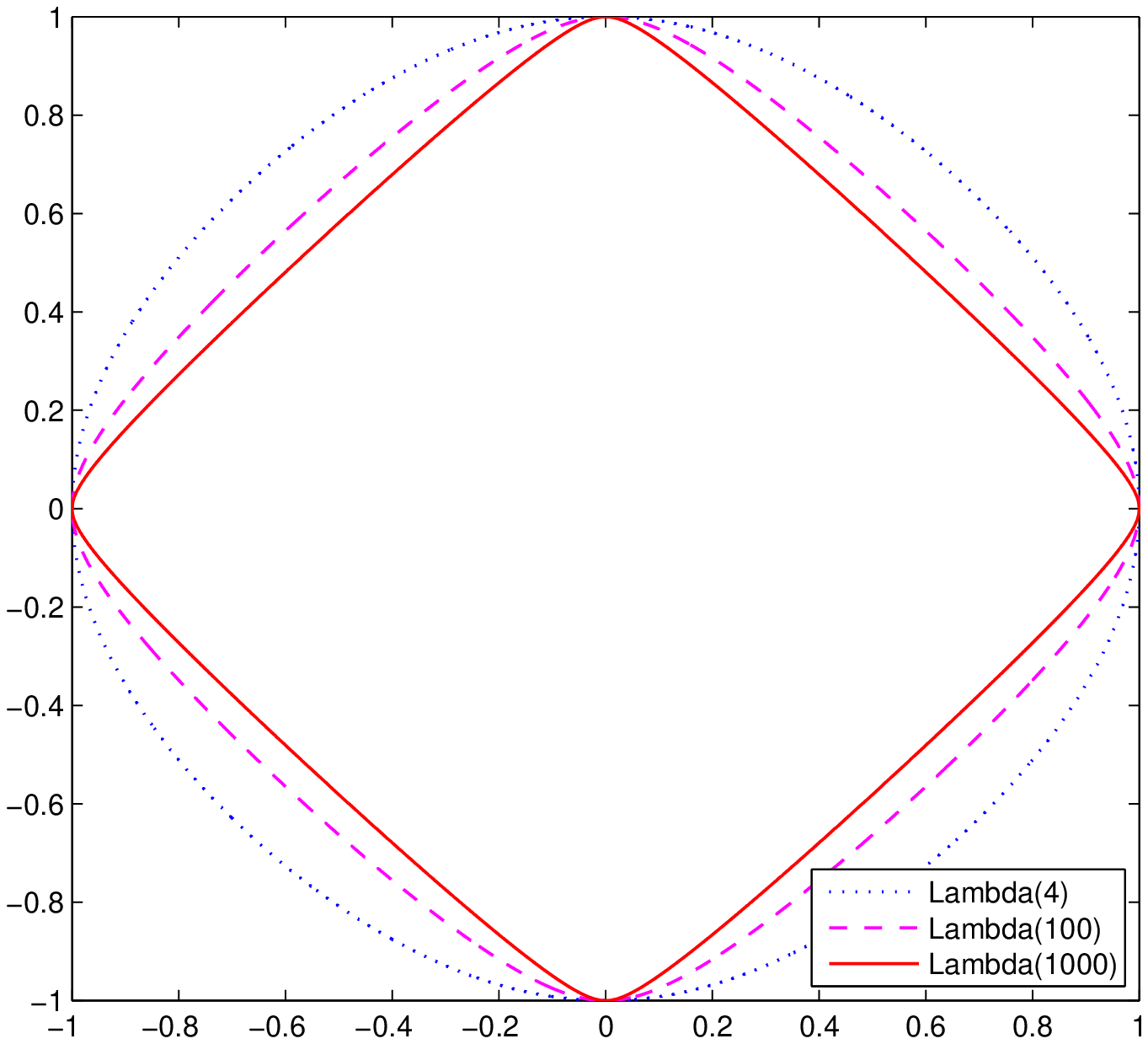}\label{fig_lapcurveset}}\hfill}
\caption{(a) $L_2$-norm (solid line) and $L_1$-norm (dashed line);
(b) KL-norm with different Laplace priors.}\label{fig_normcurves}
\end{figure*}

A more explicit illustration of the entropic regularization under a
Laplace MaxEnDNet, comparing to the conventional $L_1$ and $L_2$
regularization over an M$^3$N, can be seen in
Figure~\ref{fig_normcurves}, where the feasible regions due to the
three different norms used in the regularizer are plotted in a two
dimensional space. Specifically, it shows (1) $L_2$-norm: $w_1^2 +
w_2^2 \leq 1$; (2) $L_1$-norm: $|w_1| + |w_2| \leq 1$; and (2)
KL-norm\footnote{The curves are drawn with a symbolic computational
package to solve a equation of the form: $2x - \log{x} = a$, where
$x$ is the variable to be solved and $a$ is a constant.}:
$\sqrt{w_1^2 + 1/\lambda} + \sqrt{ w_2^2 + 1/\lambda} -
(1/\sqrt{\lambda}) \log (\sqrt{\lambda w_1^2 + 1}/2 + 1/2) -
(1/\sqrt{\lambda}) \log (\sqrt{\lambda w_1^2 + 1}/2 + 1/2) \leq b$,
where $b$ is a parameter to make the boundary pass the $(0,1)$ point
for easy comparison with the $L_2$ and $L_1$ curves. It is easy to
show that $b$ equals to $\sqrt{1/\lambda} + \sqrt{1 + 1/\lambda} -
(1/\sqrt{\lambda}) \log(\sqrt{\lambda+1}/2 + 1/2)$. It can be seen
that the $L_1$-norm boundary has sharp turning points when it passes
the axises, whereas the $L_2$ and KL-norm boundaries turn smoothly
at those points. This is the intuitive explanation of why the
$L_1$-norm directly gives sparse estimators, whereas the $L_2$-norm
and KL-norm due to a Laplace prior do not. But as shown in
Figure~\ref{fig_lapcurveset}, when the $\lambda$ gets larger and
larger, the KL-norm boundary moves closer and closer to the
$L_1$-norm boundary. When $\lambda \to \infty$, $\sqrt{w_1^2 +
1/\lambda} + \sqrt{ w_2^2 + 1/\lambda} - (1/\sqrt{\lambda}) \log
(\sqrt{\lambda w_1^2 + 1}/2 + 1/2) - (1/\sqrt{\lambda}) \log
(\sqrt{\lambda w_1^2 + 1}/2 + 1/2) \to |w_1| + |w_2|$ and $b \to 1$,
which yields exactly the $L_1$-norm in the two dimensional space.
Thus, under the linear model assumption of the discriminant
functions $F(\ \cdot \ ; \wv)$, our framework can be seen as a
smooth relaxation of the $L_1$-M$^3$N.

\section{Variational Learning of Laplace
MaxEnDNet}\label{sec:vbmethod}

Although Theorem~\ref{thrm_medn} seems to offer a general
closed-form solution to $p(\wv)$ under an arbitrary prior
$p_0(\wv)$, in practice the Lagrangian parameters $\alpha_i(\yv)$ in
$p(\wv)$ can be very hard to estimate from the dual problem D1
except for a few special choices of $p_0(\wv)$, such as a normal as
shown in Theorem~\ref{thrm_gaussianmedn}, which can be easily
generalized to any normal prior. When $p_0(\wv)$ is a Laplace prior,
as we have shown in Theorem~\ref{thrm_lapmaxendnet} and
Corollary~\ref{cor_primal_lapmaxendnet}, the corresponding dual
problem or primal problem involves a complex objective function that
is difficult to optimize. Here, we present a variational method for
an approximate learning of the Laplace MaxEnDNet.

Our approach is built on the hierarchical interpretation of the
Laplace prior as shown in Eq. (\ref{hLap}). Replacing the $p_0(\wv)$
in Problem P1 with Eq. (\ref{hLap}), and applying the Jensen's
inequality, we get an upper bound of the KL-divergence:
{\setlength\arraycolsep{1pt}\begin{eqnarray} KL(p || p_0 )
&&= -H(p) - \langle \log \int p(\wv | \tau) p(\tau | \lambda) \, \ud \tau \, \rangle_p \nonumber \\
&&\leq -H(p) - \langle \int q(\tau) \log \frac{ p(\wv | \tau) p(\tau | \lambda) } {q(\tau)} \, \ud \tau \, \rangle_p \nonumber \\
&&\triangleq \mathcal{L} (p(\wv), q(\tau)) \nonumber,
\end{eqnarray}}\\[-0.9cm]

\noindent where $q(\tauv)$ is a variational distribution used to
approximate $p(\tauv | \lambda)$. The upper bound is in fact a
KL-divergence: $\mathcal{L} (p(\wv), q(\tau)) = KL(p(\wv)q(\tau) ||
p(\wv | \tau)p(\tau | \lambda))$. Thus, $\mathcal{L}$ is convex over
$p(\wv)$, and $q(\tau)$, respectively, but not necessarily joint
convex over $(p(\wv), q(\tau))$.

Substituting this upper bound for the KL-divergence in P1, we now
solve the following Variational MaxEnDNet problem,
\begin{equation}
\hspace{-2cm} {\rm P1' ~(vMEDN)}: \quad \quad \min_{p(\wv) \in
\mathcal{F}_1; q(\tau); \xi}~\mathcal{L}(p(\wv), q(\tau)) + U(\xi).
\label{newKL}
\end{equation}

$\rm P1'$ can be solved with an iterative minimization algorithm
alternating between optimizing over $(p(\wv),\xi)$ and $q(\tauv)$,
as outlined in Algorithm~\ref{alg:vb}, and detailed below.

\begin{algorithm}[tb]
   \caption{\footnotesize Variational MaxEnDNet}
   \label{alg:vb}
\begin{algorithmic}
   \STATE {\bfseries Input:} data $\mathcal{D} = \lbrace \langle \xv^i, \yv^i \rangle \rbrace_{i=1}^N$, constants
   $C$ and $\lambda$, iteration number $T$
   \STATE {\bfseries Output:} posterior mean $\langle \wv \rangle_{p}^T$
   \STATE Initialize $\langle \wv \rangle_{p}^1 \gets 0$, $\Sigma^1 \gets I$
   \FOR{$t=1$ {\bfseries to} $T-1$}
   \STATE Step 1: solve (\ref{vmedn_dual}) or (\ref{vmedn_primal}) for $\langle \wv \rangle_{p}^{t+1} = \Sigma^t \eta$; update $\langle
   \wv \wv^\top \rangle_p^{t+1} \gets \Sigma^t + \langle \wv \rangle_p^{t+1} (\langle
   \wv \rangle_p^{t+1})^\top$.
   \STATE Step 2: use (\ref{vmedn_updatavar}) to update $\Sigma^{t+1} \gets \mathrm{diag}(\sqrt{ \frac{\langle w_k^2 \rangle_p^{t+1} }{\lambda} })$.
   \ENDFOR
\end{algorithmic}
\end{algorithm}

\textbf{Step 1:} Keep $q(\tau)$ fixed, optimize $\rm P1'$ with
respect to $(p(\wv),\xi)$. Using the same procedure as in solving P1, we get the posterior distribution $p(\wv)$ as follows,\\[-1cm]

\setlength\arraycolsep{1pt}{\begin{eqnarray} p(\wv) && \propto \exp
\lbrace \int q(\tau) \log p(\wv | \tau) \, \ud \tau \, - b \rbrace
\cdot \exp \lbrace \wv^\top \eta - \sum_{i,\yv \neq \yv^i}
\alpha_i(\yv) \Delta \ell_{i}(\yv) \rbrace \nonumber \\
&& \propto \exp \lbrace - \frac{1}{2} \wv^\top  \langle A^{-1}
\rangle_q \wv - b + \wv^\top \eta - \sum_{i,\yv \neq \yv^i}
\alpha_i(\yv) \Delta \ell_{i}(\yv) \rbrace \nonumber \\
&&= \mathcal{N}(\wv | \mu, \Sigma), \nonumber
\end{eqnarray}}\\[-0.9cm]

\noindent where $\eta = \sum_{i, \yv \neq \yv^i} \alpha_i(\yv)
\Delta \fv_{i}(\yv)$, $A = \mathrm{diag}(\tau_k)$, and $b =
KL(q(\tau) || p(\tau | \lambda))$ is a constant. The posterior mean
and variance are $\langle \wv \rangle_{p} = \mu = \Sigma \eta$ and
$\Sigma = (\langle A^{-1} \rangle_q)^{-1} = \langle \wv \wv^\top
\rangle_p - \langle \wv \rangle_p \langle \wv \rangle_p^\top$,
respectively. Note that this posterior distribution is also a normal
distribution. Analogous to the proof of Theorem 3, we can derive
that the dual parameters $\alpha$ are estimated by solving the
following dual problem:
\begin{eqnarray}\label{vmedn_dual}
&& \max_{\alpha}~\sum_{i, \yv \neq \yv^i} \alpha_i(\yv) \Delta
\ell_{i}(\yv) - \frac{1}{2} \eta^\top \Sigma \eta \\
&&\mathrm{s.t.}~~\sum_{\yv \neq \yv^i} \alpha_i(\yv) =
C;~\alpha_i(\yv) \geq 0, ~\forall i,~ \forall \yv \neq \yv^i.
\nonumber
\end{eqnarray}\\[-0.9cm]

This dual problem is now a standard quadratic program symbolically
identical to the dual of an $\mathrm{M^3N}$, and can be directly
solved using existing algorithms developed for $\mathrm{M^3N}$, such
as \citep{Taskar:03,Bartlett:04}. Alternatively, we can solve the
following primal problem:\\[-1cm]

\begin{eqnarray}\label{vmedn_primal}
&&\min_{\wv,\xi}~\frac{1}{2} \wv^\top
\Sigma^{-1} \wv + C \sum_{i=1}^N \xi_i \\
&&\mathrm{s.t.}~~\wv^\top \Delta \mathrm{\mathbf{f}}_i(\yv) \geq
\Delta \ell_{i}(\yv) - \xi_i;~\xi_i \geq 0,~~\forall i,~\forall
\yv \neq \yv^i. \nonumber
\end{eqnarray}\\[-1.0cm]

Based on the proof of Corollary~\ref{cor_primal_gaussianmedn}, it is
easy to show that the solution of the problem~(\ref{vmedn_primal})
leads to the posterior mean of $\wv$ under $p(\wv)$, which will be
used to do prediction by $h_1$. The primal problem can be solved
with the subgradient \citep{Ratliff:07},
cutting-plane~\citep{Tsochantaridis:04}, or extragradient
\citep{Taskar:06} method.

\textbf{Step 2}: Keep $p(\wv)$ fixed, optimize ${\rm P1'}$ with
respect to $q(\tau)$. Taking the derivative of $\mathcal{L}$ with
respect to $q(\tau)$ and set it to zero, we get:
\setlength\arraycolsep{1pt}{\begin{eqnarray} q(\tau) &&\propto
p(\tau | \lambda) \exp \big \lbrace \langle \log p(\wv |\tau)
\rangle_p \big \rbrace. \nonumber
\end{eqnarray}}\\[-0.9cm]

\noindent Since both $p(\wv | \tau)$ and $p(\tau | \lambda)$ can be
written as a product of univariate Gaussian and univariate
exponential distributions, respectively, over each dimension,
$q(\tau)$ also factorizes over each dimension: $q(\tau) =
\prod_{k=1}^K q(\tau_k)$, where each $q(\tau_k)$ can be expressed
as: \setlength\arraycolsep{1pt}{\begin{eqnarray} \forall k: \quad
q(\tau_k) &&\propto p(\tau_k | \lambda) \exp \big \lbrace \langle
\log p(w_k|\tau_k) \rangle_p \big \rbrace \nonumber \\
&&\propto \mathcal{N}( \sqrt{ \langle w_k^2 \rangle_p } | 0,
\tau_k) \exp(- \frac{1}{2} \lambda \tau_k). \nonumber
\end{eqnarray}}\\[-0.9cm]

\noindent The same distribution has been derived in
\citep{Kaban:07}, and similar to the hierarchical representation of
a Laplace distribution we can get the normalization factor: $\int
\mathcal{N}( \sqrt{ \langle w_k^2 \rangle_p } | 0, \tau_k) \cdot
\frac{\lambda}{2} \exp(- \frac{1}{2} \lambda \tau_k) \, \ud \tau_k
\, = \frac{\sqrt{\lambda}}{2} \exp ( -\sqrt{\lambda \langle w_k^2
\rangle_p } )$. Also, as in \citep{Kaban:07}, we can calculate the
expectations $\langle \tau_k^{-1} \rangle_q$ which are required in
calculating $\langle A^{-1} \rangle_q$ as follows,
\begin{equation}\label{vmedn_updatavar}
\langle \frac{1}{\tau_k} \rangle_q = \int
\frac{1}{\tau_k} q(\tau_k) \, \ud \tau_k \, = \sqrt{
\frac{\lambda}{\langle w_k^2 \rangle_p } }.
\end{equation}

We iterate between the above two steps until convergence. Due to the
convexity (not joint convexity) of the upper bound, the algorithm is
guaranteed to converge to a local optimum. Then, we apply the
posterior distribution $p(\wv)$, which is in the form of a normal
distribution, to make prediction using the averaging prediction law
in Eq. (\ref{BM3N}). Due to the shrinkage effect of the Laplacian
entropic regularization discussed in
Section~\ref{sec:shrinkageanalysis}, for irrelevant features, the
variances should converge to zeros and thus lead to a sparse
estimation of $\wv$. To summarize, the intuition behind this
iterative minimization algorithm is as follows. First, we use a
Gaussian distribution to approximate the Laplace distribution and
thus get a QP problem that is analogous to that of the standard
M$^3$N; then, in the second step we update the covariance matrix in
the QP problem with an exponential hyper-prior on the variance.

\section{Generalization Bound}

The PAC-Bayes theory for averaging classifiers \citep{Langford:01}
provides a theoretical motivation to learn an averaging model for
classification. In this section, we extend the classic PAC-Bayes
theory on binary classifiers to MaxEnDNet, and analyze the
generalization performance of the structured prediction rule $h_1$
in Eq. (\ref{BM3N}). In order to prove an error bound for $h_1$, the
following mild assumption on the boundedness of discriminant
function $F(\ \cdot \ ; \wv)$ is necessary, i.e., there exists a
positive constant $c$, such that,
\[ \hspace{-1cm} \forall \wv, \quad
F(\ \cdot \ ; \wv) \in \mathcal{H}: \quad \mathcal{X} \times
\mathcal{Y} \to \lbrack -c, c \rbrack.
\]
Recall that the averaging structured prediction function under the
MaxEnDNet is defined as $h(\xv, \yv) = \langle F(\xv, \yv; \wv)
\rangle_{p(\wv)}$. Let's define the predictive margin of an instance
$(\xv, \yv)$ under a function $h$ as $M(h, \xv, \yv) = h(\xv, \yv) -
\max_{\yv^\prime \neq \yv} h(\xv, \yv^\prime)$. Clearly, $h$ makes a
wrong prediction on $(\xv, \yv)$ only if $M(h, \xv, \yv) \leq 0$.
Let $Q$ denote a distribution over $\mathcal{X} \times \mathcal{Y}$,
and let $\mathcal{D}$ represent a sample of $N$ instances randomly
drawn from $Q$. With these definitions, we have the following
structured version of PAC-Bayes theorem.

\begin{theorem}[PAC-Bayes Bound of MaxEnDNet]\label{thrm_pacbayes} Let $p_0$ be any continuous
probability distribution over $\mathcal{H}$ and let $\delta \in (0,
1)$. If $F(\ \cdot \ ; \wv) \in \mathcal{H}$ is bounded by $\pm c$
as above, then with probability at least $1 - \delta$, for a random
sample $\mathcal{D}$ of $N$ instances from $Q$, for every
distribution $p$ over $\mathcal{H}$, and for all margin thresholds
$\gamma > 0$:\\[-0.9cm]

\setlength\arraycolsep{1pt} {\begin{eqnarray} \mathrm{Pr}_{Q} ( M(h,
\xv, \yv ) \leq 0 ) \leq \mathrm{Pr}_{\mathcal{D}} ( M(h, \xv, \yv)
\leq \gamma ) + O \Big( \sqrt{ \frac{\gamma^{-2} KL(p||p_0) \ln (N
|\mathcal{Y}|) + \ln N + \ln \delta^{-1}}{N} } \Big), \nonumber
\end{eqnarray}}\\[-0.9cm]

\noindent where $\mathrm{Pr}_{Q}(.)$ and
$\mathrm{Pr}_{\mathcal{D}}(.)$ represent the probabilities of events
over the true distribution $Q$, and over the empirical distribution
of $\mathcal{D}$, respectively.
\end{theorem}

The proof of Theorem \ref{thrm_pacbayes} follows the same spirit of
the proof of the original PAC-Bayes bound, but with a number of
technical extensions dealing with structured outputs and margins.
See appendix B.5 for the details.

Recently, \cite{McAllester:07} presents a {\it stochastic}
max-margin structured prediction model, which is different from the
averaging predictor under the MaxEnDNet model, by designing a
``posterior" distribution from which a model is sampled to make
prediction. A PAC-Bayes bound with an improved dependence on
$|\mathcal{Y}|$ was shown in this model. \cite{Langford:03} show an
interesting connection between the PAC-Bayes bounds for averaging
classifiers and stochastic classifiers, again by designing a
posterior distribution. But our posterior distribution is solved
with MaxEnDNet and is generally different from those designed in
\citep{McAllester:07} and \citep{Langford:03}.

\section{Experiments}

In this section, we present empirical evaluations of the proposed
Laplace MaxEnDNet (LapMEDN) on both synthetic and real data sets. We
compare LapMEDN with M$^3$N (i.e., the Gaussian MaxEnDNet),
$L_1$-regularized M$^3$N ($L_1$-M$^3$N), CRFs, $L_1$-regularized
CRFs ($L_1$-CRFs), and $L_2$-regularized CRFs ($L_2$-CRFs). We use
the quasi-Newton method \citep{Liu:89} and its variant
\citep{Andrew:07} to solve the optimization problem of CRFs,
$L_1$-CRFs, and $L_2$-CRFs. For M$^3$N and LapMEDN,
we use the sub-gradient method \citep{Ratliff:07} to solve the
corresponding primal problem. To the best of our knowledge, no
formal description, implementation, and evaluation of the
$L_1$-M$^3$N exist in the literature, therefore how to solve
$L_1$-M$^3$N remains an open problem and for comparison purpose we
had to develop this model and algorithm anew. Details of our work
along this line deserves a more thorough presentation, which is
beyond the scope of this paper and will appear elsewhere. But
briefly, for our experiments on synthetic data, we implemented the
constraint generating method~\citep{Tsochantaridis:04} which uses
MOSEK to solve an equivalent LP re-formulation of $L_1$-M$^3$N.
However, this approach is extremely slow on larger problems;
therefore on real data we instead applied the sub-gradient method
\citep{Ratliff:07} with a projection to an
$L_1$-ball~\citep{Duchi:08} to solve the larger $L_1$-M$^3$N based
on the equivalent re-formulation with an $L_1$-norm constraint
(i.e., the second formulation in Appendix A).

\subsection{Evaluation on Synthetic Data}

We first evaluate all the competing models on synthetic data where
the true structured predictions are known. Here, we consider
sequence data, i.e., each input $\xv$ is a sequence $(x_1, \dots,
x_L)$, and each component $x_l$ is a $d$-dimensional vector of input
features. The synthetic data are generated from pre-specified
conditional random field models with either i.i.d. instantiations of
the input features (i.e., elements in the $d$-dimensional feature
vectors) or correlated (i.e., structured) instantiations of the
input features, from which samples of the structured output $\yv$,
i.e., a sequence $(y_1, \dots, y_L)$, can be drawn from the
conditional distribution $p(\yv|\xv)$ defined by the CRF based on a
Gibbs sampler.


\subsubsection{I.i.d. input features}

The first experiment is conducted on synthetic sequence data with
100 i.i.d. input features (i.e., $d=100$). We generate three types
of data sets with 10, 30, and 50 relevant input features,
respectively. For each type, we randomly generate 10 linear-chain
CRFs with 8 binary labeling states (i.e., $L=8$ and $\mathcal{Y}_l =
\{0, 1\}$). The feature functions include: a real valued
state-feature function over a one dimensional input feature and a
class label; and 4 ($2\times2$) binary transition feature functions
capturing pairwise label dependencies. For each model we generate a
data set of 1000 samples. For each sample, we first {\it
independently} draw the 100 input features from a standard normal
distribution, and then apply a Gibbs sampler (based on the
conditional distribution of the generated CRFs) to assign a labeling
sequence with 5000 iterations.

For each data set, we randomly draw a subset as training data and
use the rest for testing. The sizes of training set are 30, 50, 80,
100, and 150. The QP problem in M$^3$N and the first step of LapMEDN
is solved with the exponentiated gradient method
\citep{Bartlett:04}. In all the following experiments, the
regularization constants of $L_1$-CRFs and $L_2$-CRFs are chosen
from $\lbrace 0.01, 0.1, 1, 4, 9, 16\rbrace$ by a 5-fold
cross-validation during the training. For the LapMEDN, we use the
same method to choose $\lambda$ from 20 roughly evenly spaced values
between 1 and 268. For each setting, a performance score is computed
from the average over 10 random samples of data sets.

\begin{figure}
\centerline{\includegraphics[width=480pt]{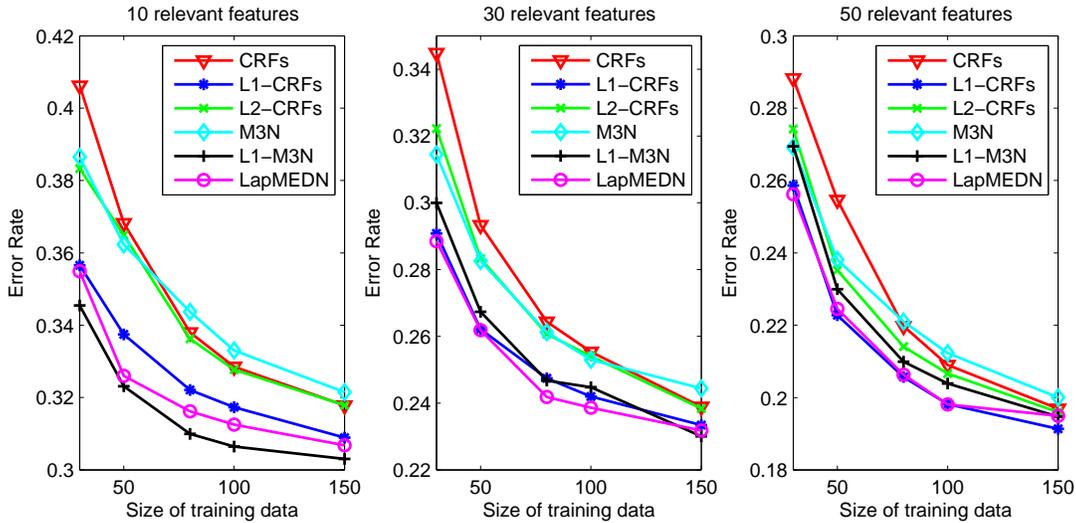}}%
\vspace{-0.4cm} \caption{Evaluation results on data sets with
i.i.d features.} \label{fig_resiid} 
\end{figure}

The results are shown in Figure~\ref{fig_resiid}. All the results of
the LapMEDN are achieved with 3 iterations of the variational
learning algorithm. From the results, we can see that under
different settings LapMEDN consistently outperforms M$^3$N and
performs comparably with $L_1$-CRFs and $L_1$-M$^3$N, both of which
encourage a sparse estimate; and both the $L_1$-CRFs and $L_2$-CRFs
outperform the un-regularized CRFs, especially in the cases where
the number of training data is small. One interesting result is that
the M$^3$N and $L_2$-CRFs perform comparably. This is reasonable
because as derived by \cite{Lebanon:01} and noted by
\cite{Globerson:07} that the $L_2$-regularized maximum likelihood
estimation of CRFs has a similar convex dual as that of the M$^3$N,
and the only difference is the loss they try to optimize, i.e., CRFs
optimize the log-loss while M$^3$N optimizes the hinge-loss. Another
interesting observation is that when there are very few relevant
features, $L_1$-M$^3$N performs the best (slightly better than
LapMEDN); but as the number of relevant features increases LapMEDN
performs slightly better than the $L_1$-M$^3$N. Finally, as the
number of training data increases, all the algorithms consistently
achieve better performance.

\begin{figure} \centerline{ \includegraphics[width=260pt]{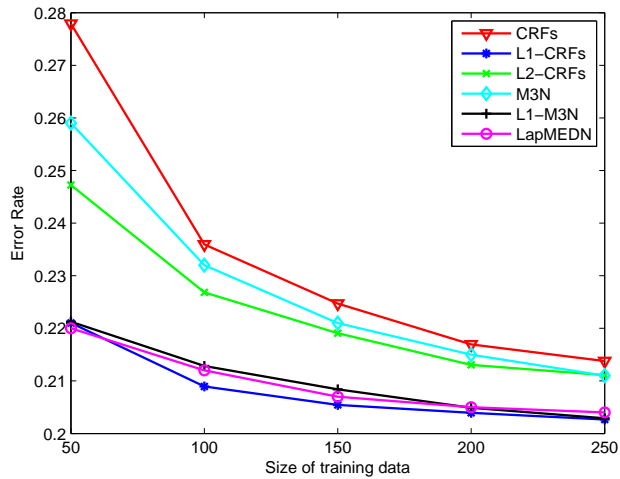}
\label{fig_relvres}} \vspace{-0.4cm} \caption{Results on data sets
with 30 relevant features.} \label{fig_relvRes}
\end{figure}

\subsubsection{Correlated input features}

In reality, most data sets contain redundancies and the input
features are usually correlated. So, we evaluate our models on
synthetic data sets with correlated input features. We take the
similar procedure as in generating the data sets with i.i.d.
features to first generate 10 linear-chain CRF models. Then, each
CRF is used to generate a data set that contain 1000 instances, each
with 100 input features of which 30 are relevant to the output. The
30 relevant input features are partitioned into 10 groups. For the
features in each group, we first draw a real-value from a standard
normal distribution and then corrupt the feature with a random
Gaussian noise to get 3 correlated features. The noise Gaussian has
a zero mean and standard variance 0.05. Here and in all the
remaining experiments, we use the sub-gradient method
\citep{Ratliff:07} to solve the QP problem in both M$^3$N and the
variational learning algorithm of LapMEDN. We use the learning rate
and complexity constant that are suggested by the authors, that is,
$\alpha_t = \frac{1}{2\beta \sqrt{t}}$~and~$C = 200\beta$, where
$\beta$ is a parameter we introduced to adjust $\alpha_t$ and $C$.
We do K-fold CV on each data set and take the average over the 10
data sets as the final results. Like \citep{Taskar:03}, in each run
we choose one part to do training and test on the rest K-1 parts. We
vary K from 20, 10, 7, 5, to 4. In other words, we use 50, 100,
about 150, 200, and 250 samples during the training. We use the same
grid search to choose~$\lambda$~and~$\beta$ from $\lbrace 9, 16, 25,
36, 49, 64\rbrace$ and $\lbrace 1, 10, 20, 30, 40, 50, 60\rbrace$
respectively. Results are shown in Figure~\ref{fig_relvRes}. We can
get the same conclusions as in the previous results.

\begin{figure}
\centerline{\includegraphics[width=440pt]{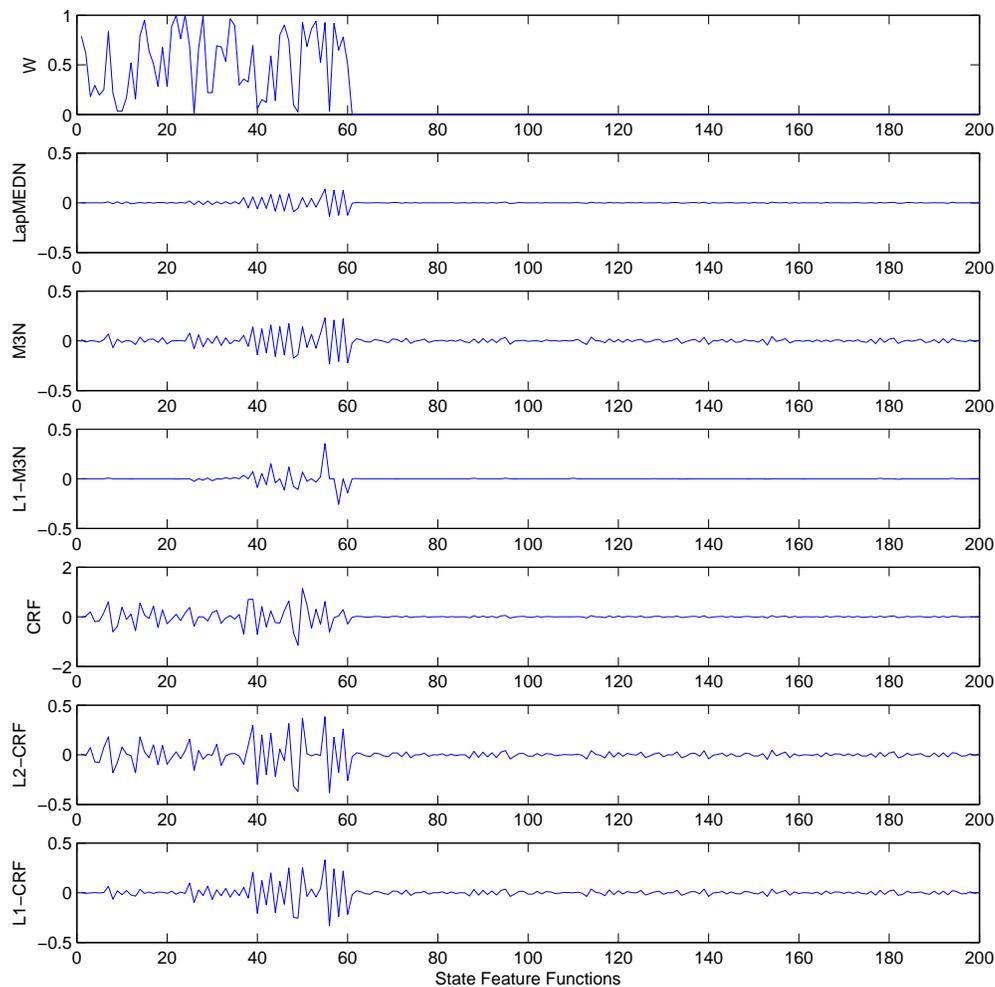}}
\vspace{-0.8cm}\caption{From top to bottom, plot 1 shows the weights
of the state feature functions in the linear-chain CRF model from
which the data are generated; plot 2 to plot 7 show the average
weights of the learned LapMEDN, M$^3$N, $L_1$-M$^3$N, CRFs,
$L_2$-CRFs, and $L_1$-CRFs over 10 fold CV,
respectively.}\label{fig_avgVar}
\end{figure}

\begin{figure}
\centerline{ \includegraphics[width=440pt]{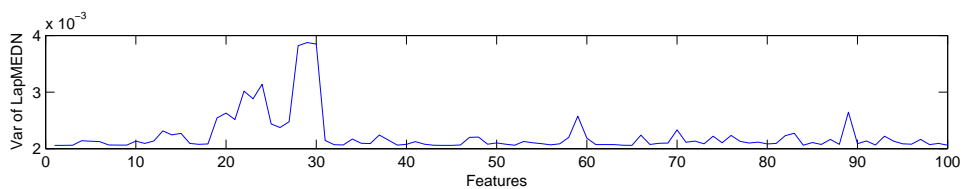}} \caption{The
average variances of the features on the first data set by
LapMEDN.}\label{fig_var}
\end{figure}

Figure~\ref{fig_avgVar}~shows the true weights of the corresponding
200 state feature functions in the model that generates the first
data set, and the average of estimated weights of these features
under all competing models fitted from the first data set. All the
averages are taken over 10 fold cross-validation. From the plots (2
to 7) of the average model weights, we can see that: for the last
140 state feature functions, which correspond to the last 70
irrelevant features, their average weights under LapMEDN (averaged
posterior means $\wv$ in this case), $L_1$-M$^3$N and $L_1$-CRFs are
extremely small, while CRFs and $L_2$-CRFs can have larger values;
for the first 60 state feature functions, which correspond to the 30
relevant features, the overall weight estimation under LapMEDN is
similar to that of the sparse $L_1$-CRFs and $L_1$-M$^3$N, but
appear to exhibit more shrinkage. Noticeably, CRFs and $L_2$-CRFs
both have more feature functions with large average weights. Note
that all the models have quite different average weights from the
model (see the first plot) that generates the data. This is because
we use a stochastic procedure (i.e., Gibbs sampler) to assign labels
to the generated data samples instead of using the labels that are
predicted by the model that generates the data. In fact, if we use
the model that generates the data to do prediction on its generated
data, the error rate is about 0.5. Thus, the learned models, which
get lower error rates, are different from the model that generates
the data. Figure~\ref{fig_var}~shows the variances of the
100-dimensional input features (since the variances of the two
feature functions that correspond to the same input feature are the
same, we collapse each pair into one point) learned by LapMEDN.
Again, the variances are the averages over 10 fold cross-validation.
From the plot, we can see that the LapMEDN 
can recover the correlation among the features to some extend, e.g.,
for the first 30 correlated features, which are the relevant to the
output, the features in the same group tend to have similar
(average) variances in LapMEDN, whereas there is no such correlation
among all the other features. From these observations in both
Figure~\ref{fig_avgVar} and ~\ref{fig_var}, we can conclude that
LapMEDN can reasonably recover the sparse structures in the input
data.

\subsection{Real-World OCR Data Set}

The OCR data set is partitioned into 10 subsets for 10-fold CV as in
\citep{Taskar:03,Ratliff:07}. We randomly select $N$ samples from
each fold and put them together to do 10-fold CV. We vary $N$ from
100, 150, 200, to 250, and denote the selected data sets by OCR100,
OCR150, OCR200, and OCR250, respectively. On these data sets and the
web data as in Section~\ref{sec_webdata}, our implementation of the
cutting plane method for $L_1$-M$^3$N is extremely slow. The
warm-start simplex method of MOSEK does not help either. For
example, if we stop the algorithm with 600 iterations on OCR100,
then it will take about 20 hours to finish the 10 fold CV. Even with
more than 5 thousands of constraints in each training, the
performance is still very bad (the error rate is about 0.45). Thus,
we turn to an approximate projected sub-gradient method to solve the
$L_1$-M$^3$N by combining the on-line subgradient
method~\citep{Ratliff:07} and the efficient $L_1$-ball projection
algorithm~\citep{Duchi:08}. The projected sub-gradient method does
not work so well as the cutting plane method on the synthetic data
sets. That's why we use two different methods.

\begin{figure}
\centerline{\includegraphics[width=480pt]{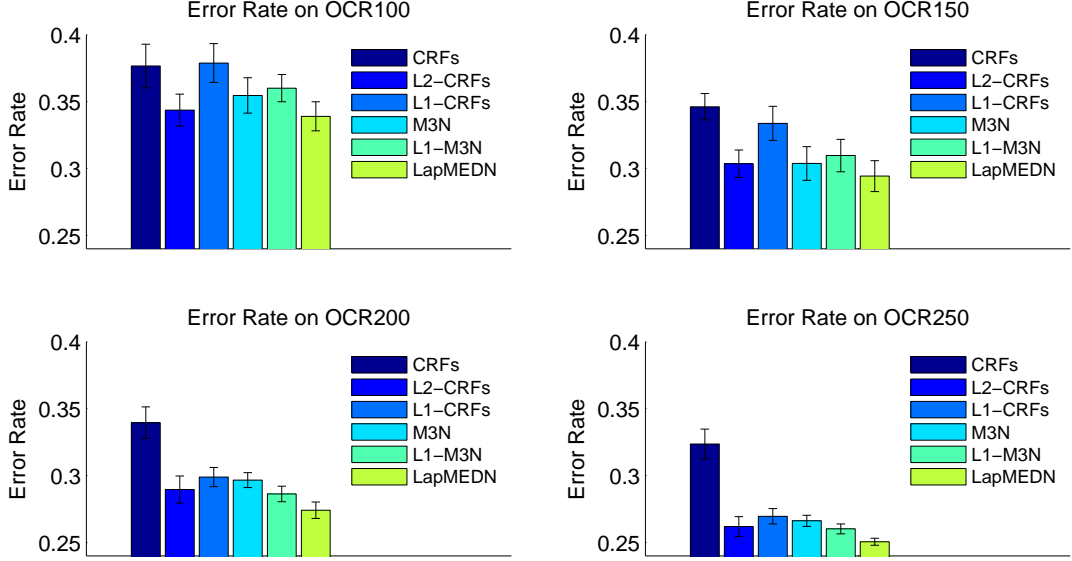}}%
\vspace{-0.4cm} \caption{Evaluation results on OCR data set with different numbers of selected data.} \label{fig_ocrresbar}
\end{figure}

\begin{figure*}%
\centerline{\includegraphics[width=440pt]{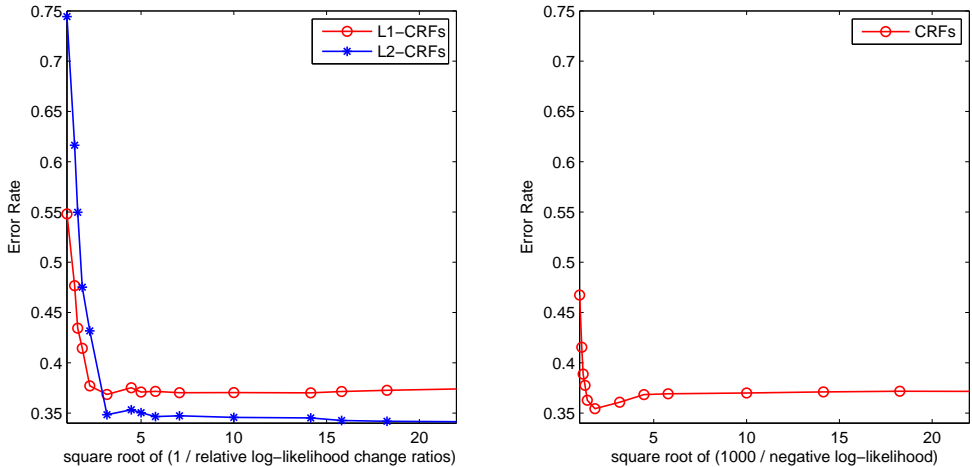}}%
\caption{The error rates of CRF models on test data during the
learning. For the left plot, the horizontal axis is $\sqrt{1 /
ratioLL}$, where $ratioLL$ is the relative change ratios of the
log-likelihood and from left to right, the change ratios are 1, 0.5,
0.4, 0.3, 0.2, 0.1, 0.05, 0.04, 0.03, 0.02, 0.01, 0.005, 0.004,
0.003, 0.002, 0.001, 0.0005, 0.0004, 0.0003, 0.0002, 0.0001, and
0.00005; for the right plot, the horizontal axis is $\sqrt{1000 /
negLL}$, where $negLL$ is the negative log-likelihood, and from left
to right $negLL$ are 1000, 800, 700, 600, 500, 300, 100, 50, 30, 10,
5, 3, 1, 0.5, 0.3, 0.1, 0.05, 0.03, 0.01, 0.005, 0.003, and 0.002.}
\label{fig_overfitting}
\end{figure*}

For $\beta = 4$ on OCR100 and OCR150, $\beta = 2$ on OCR200 and
OCR250, and $\lambda=36$, the results are shown in
Figure~\ref{fig_ocrresbar}. We can see that as the number of
training instances increases, all the algorithms get lower error
rates and smaller variances. Generally, the LapMEDN consistently
outperforms all the other models. M$^3$N outperforms the standard,
non-regularized, CRFs and the $L_1$-CRFs. Again, $L_2$-CRFs perform
comparably with M$^3$N. This is a bit surprising but still
reasonable due to the understanding of their only difference on the
loss functions~\citep{Globerson:07} as we have stated. By examining
the prediction accuracy during the learning, we can see an obvious
over-fitting in CRFs and $L_1$-CRFs as shown in
Figure~\ref{fig_overfitting}. In contrast, $L_2$-CRFs are very
robust. This is because unlike the synthetic data sets, features in
real-world data are usually not completely irrelevant. In this case,
putting small weights to zero as in $L_1$-CRFs will hurt
generalization ability and also lead to instability to
regularization constants as shown later. Instead, $L_2$-CRFs do not
put small weights to zero but shrink them towards zero as in the
LapMEDN. The non-regularized maximum likelihood estimation can
easily lead to over-fitting too. For the two sparse models, the
results suggest the potential advantages of $L_1$-norm regularized
M$^3$N, which are consistently better than the $L_1$-CRFs.
Furthermore, as we shall see later, $L_1$-M$^3$N is more stable than
the $L_1$-CRFs.

\begin{figure*}[tb]%
\centerline{\subfigure{\includegraphics[width=440pt]{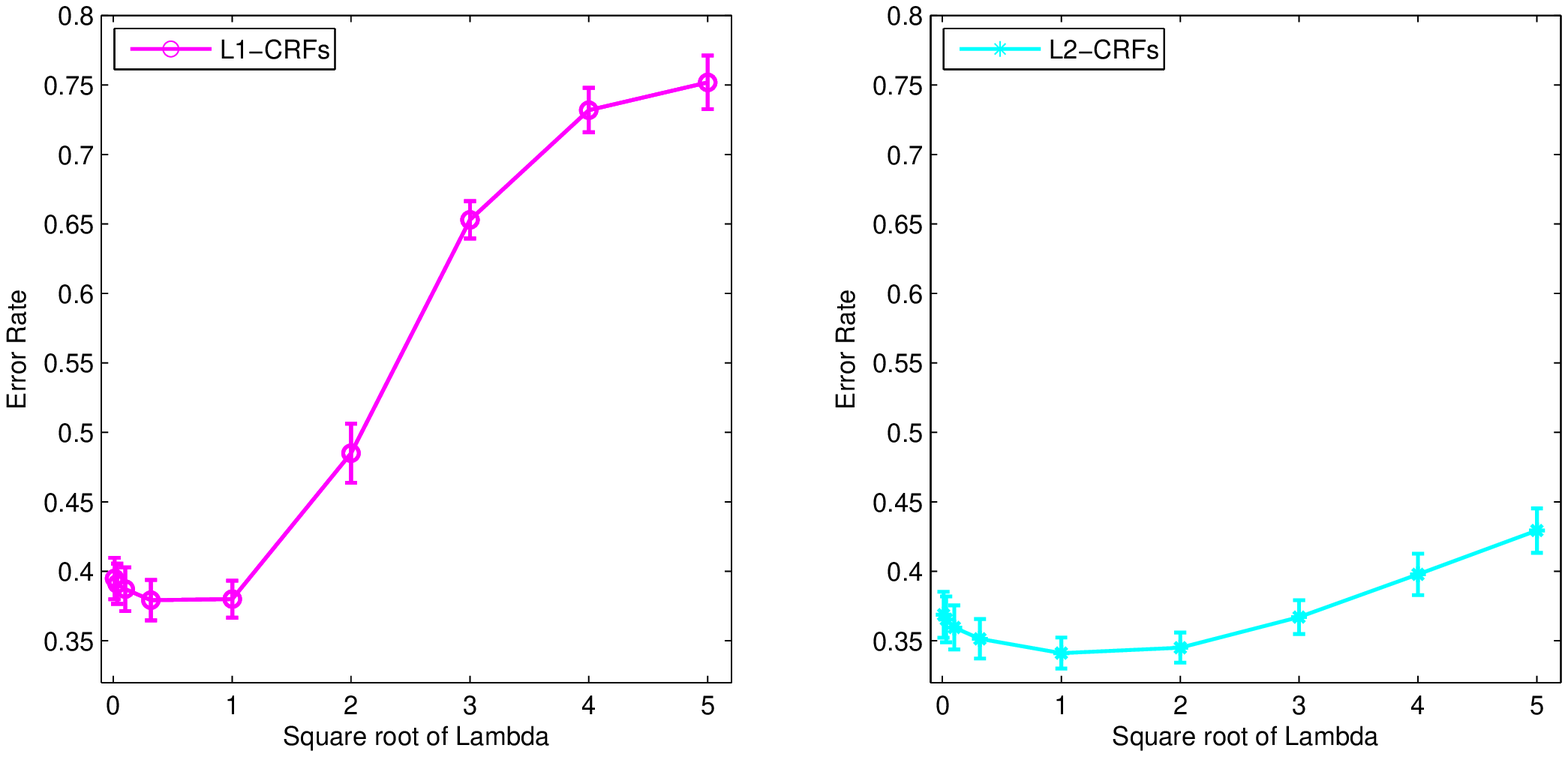}}\label{sensitivity_crf}}%
\vspace{-0.4cm}
\centerline{\subfigure{\includegraphics[width=440pt]{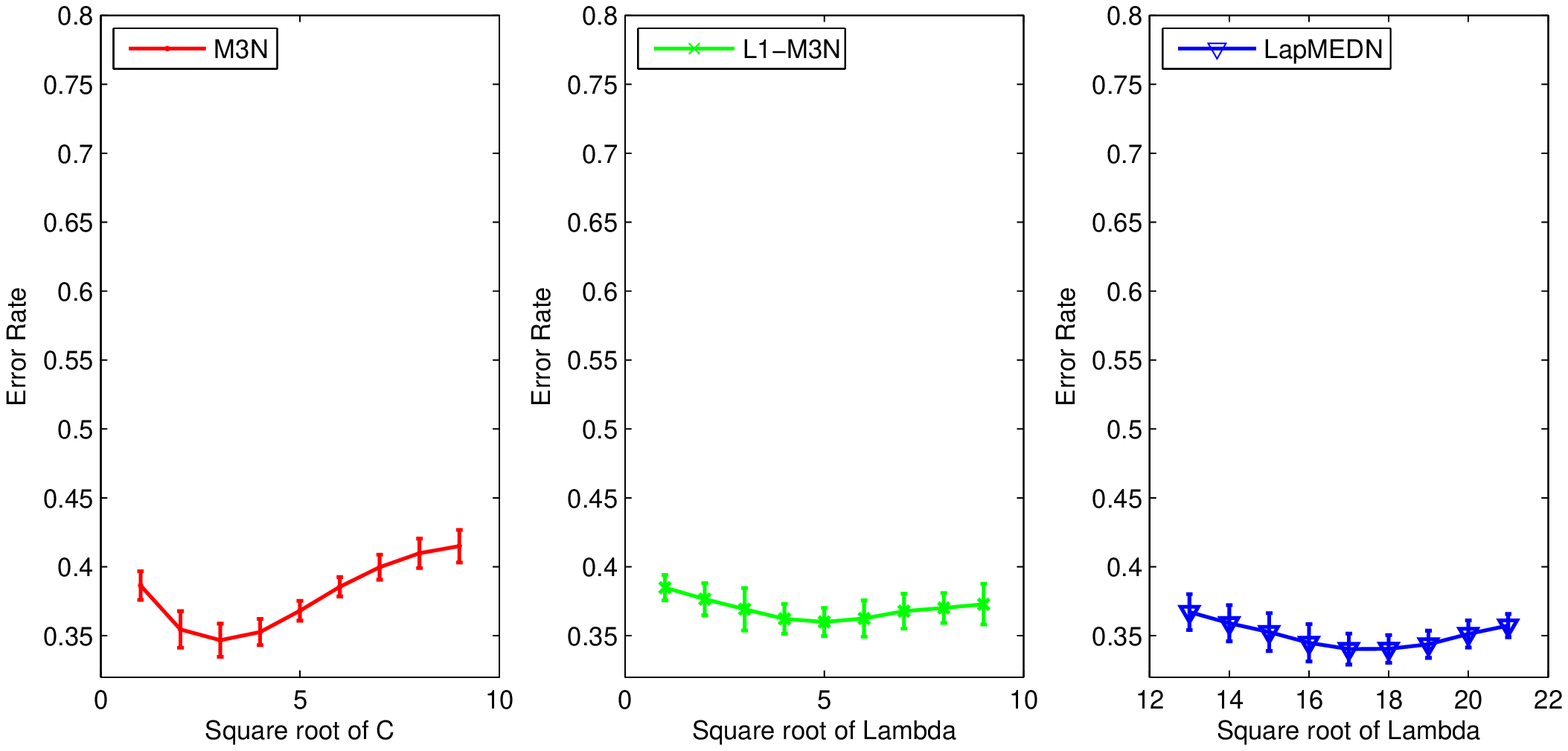}}\label{sensitivity_m3n}}%
\caption{Error rates of different models on OCR100 with different
regularization constants. The regularization constant is the
parameter $C$ for M$^3$N, and for all the other models, it is the
parameter $\lambda$. From left to right, the regularization
constants for the two regularized CRFs (above plot) are 0.0001,
0.001, 0.01, 0.1, 1, 4, 9, 16, and 25; for $M^3$N and LapMEDN, the
regularization constants are $k^2,~1 \leq k \leq 9$; and for
$L_1$-M$^3$N, the constants are $k^2,~13\leq k \leq
21$.}\label{sensitivity}
\end{figure*}

\subsection{Sensitivity to Regularization Constants}

Figure~\ref{sensitivity}~shows the error rates of the models in
question on the data set OCR100 over different magnitudes of the
regularization constants. For M$^3$N, the regularization constant is
the parameter $C$, and for all the other models, the regularization
constant is the parameter $\lambda$. When the $\lambda$ changes, the
parameter $C$ in LapMEDN and $L_1$-M$^3$N is fixed at the unit 1.

From the results, we can see that the $L_1$-CRFs are quite sensitive
to the regularization constants. However, $L_2$-CRFs,
$\mathrm{M^3N}$, $L_1$-M$^3$N and LapMEDN are much less sensitive.
LapMEDN and $L_1$-M$^3$N are the most stable models. The stability
of LapMEDN is due to the posterior weighting instead of
hard-thresholding to set small weights to zero as in the $L_1$-CRFs.
One interesting observation is that the max-margin based
$L_1$-M$^3$N is much more stable compared to the $L_1$-norm
regularized CRFs. One possible reason is that like LapMEDN,
$L_1$-M$^3$N enjoys both the primal and dual sparsity, which makes
it less sensitive to outliers; whereas the $L_1$-CRF is only primal
sparse.

\subsection{Real-World Web Data Extraction}\label{sec_webdata}

The last experiments are conducted on another problem regarding the
real world web data extraction, as extensively studied in
\citep{Zhu:08c}. Web data extraction is a task to identify
interested information from web pages. Each sample is a data record
or an entire web page which is represented as a set of HTML
elements. One striking characteristic of web data extraction is that
various types of structural dependencies between HTML elements
exist, e.g. the HTML tag tree or the Document Object Model (DOM)
structure is itself hierarchical. In \citep{Zhu:08c}, hierarchical
CRFs are shown to have great promise and achieve better performance
than flat models like linear-chain CRFs~\citep{Lafferty:01}. One
method to construct a hierarchical model is to first use a parser to
construct a so called vision tree. Then, based on the vision tree, a
hierarchical model can be constructed accordingly to extract the
interested attributes, e.g. a product's name, image, price,
description, etc. See \citep{Zhu:08c} for an example of the vision
tree and the corresponding hierarchical model. In such a
hierarchical extraction model, inner nodes are useful to incorporate
long distance dependencies, and the variables at one level are
refinements of the variables at upper levels.

In these experiments\footnote{These experiments are slightly
different from those in \citep{Zhu:08c}. Here, we introduce more
general feature functions based on the content and visual features
as in \citep{Zhu:08c}.}, we identify product items for sale on the
Web. For each product item, four attributes---{\it Name}, {\it
Image}, {\it Price}, and {\it Description} are extracted. We use the
data set that is built with web pages generated by 37 different
templates \citep{Zhu:08c}. For each template, there are 5 pages for
training and 10 for testing. We evaluate all the methods on the {\it
record level}, that is, we assume that data records are given, and
we compare different models on the accuracy of extracting attributes
in the given records. In the 185 training pages, there are 1585 data
records in total; in the 370 testing pages, 3391 data records are
collected. As for the evaluation criteria, we use the two
comprehensive measures, i.e. average F1 and block instance accuracy.
As defined in~\citep{Zhu:08c}, average F1 is the average value of
the F1 scores of the four attributes, and block instance accuracy is
the percent of data records whose {\it Name}, {\it Image}, and {\it
Price} are all correctly identified.

\begin{figure*}%
\centerline{\includegraphics[width=440pt]{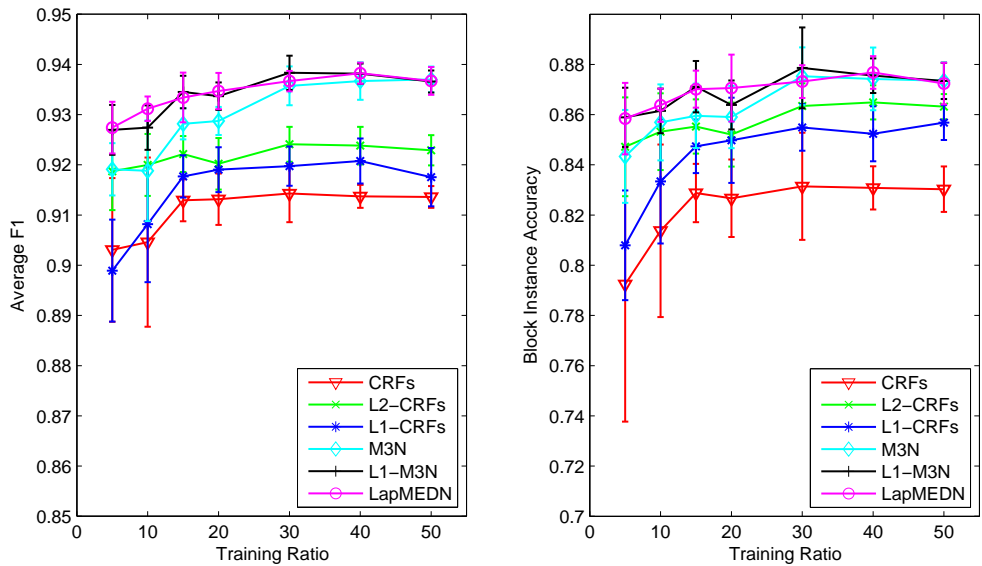}}%
\caption{The average F1 values and block instance accuracy on web
data extraction with different number of training data.}
\label{fig_webdata}
\end{figure*}

We randomly select $m= 5, 10, 15, 20, 30, 40, {\rm or}, 50$ percent
of the training records as training data, and test on all the
testing records. For each $m$, 10 independent experiments were
conducted and the average performance is summarized in
Figure~\ref{fig_webdata}. From the results, we can see that all:
first, the models (especially the max-margin models, i.e., M$^3$N,
$L_1$-M$^3$N, and LapMEDN) with regularization (i.e., $L_1$-norm,
$L_2$-norm, or the entropic regularization of LapMEDN) can
significantly outperform the un-regularized CRFs. Second, the
max-margin models generally outperform the conditional
likelihood-based models (i.e., CRFs, $L_2$-CRFs, and $L_1$-CRFs).
Third, the LapMEDN perform comparably with the $L_1$-M$^3$N, which
enjoys both dual and primal sparsity as the LapMEDN, and outperforms
all other models, especially when the number of training data is
small. Finally, as in the previous experiments on OCR data, the
$L_1$-M$^3$N generally outperforms the $L_1$-CRFs, which suggests
the potential promise of the max-margin based $L_1$-M$^3$N. A
detailed discussion and validation this new model ($L_1$-M$^3$N) is
beyond the scope of this paper, and will be deferred to a later
paper.

\section{Related Work}

Our work is motivated by the maximum entropy discrimination (MED)
method proposed by \citep{Jaakkola:99}, which integrates SVM and
entropic regularization to obtain an averaging maximum margin model
for classification. The MaxEnDNet model presented is essentially a
structured version of MED built on M$^3$N---the so called
``structured SVM". As we presented in this paper, this extension
leads to a substantially more flexible and powerful new paradigm for
structured discriminative learning and prediction, which enjoys a
number of advantages such as model averaging, primal and dual
sparsity, accommodation of latent generative structures, but at the
same time as raises new algorithmic challenges in inference and
learning.

Related to our approach, a sparse Bayesian learning framework has
been proposed to find sparse and robust solutions to regression and
classification. One example along this line is the relevance vector
machine (RVM) \citep{Tipping:01}. The RVM was proposed based on SVM.
But unlike SVM which directly optimizes on the margins, RVM defines
a likelihood function from the margins with a Gaussian distribution
for regression and a logistic sigmoid link function for
classification and then does \textit{type-II maximum likelihood}
estimation, that is, RVM maximizes the \textit{marginal likelihood}.
Although called {\it sparse Bayesian learning}
\citep{Figueiredo:01a,Eyheramendy:03}, as shown in \citep{Kaban:07}
the sparsity is actually due to the MAP estimation. The similar
ambiguity of RVM is justified in \citep{Wipf:03}. Unlike these
approaches, we adhere to a full Bayesian-style principle and learn a
distribution of predictive models by optimizing a generalized
maximum entropy under a set of the \textit{expected} margin
constraints. By defining likelihood functions with margins, similar
Bayesian interpretations of both binary and multi-class SVM can also
be found in \citep{Sollich:02,Zhang:06}.

The hierarchical interpretation of the Laplace prior has been
explored in a number of contexts in the literature. Based on this
interpretation, a Jeffrey's non-informative second-level hyper-prior
was proposed in \citep{Figueiredo:01a}, with an EM algorithm
developed to find the MAP estimate. The advantage of the Jeffrey's
prior is that it is parameter-free. But as shown in
\citep{Eyheramendy:03,Kaban:07}, usually no advantage is achieved by
using the Jeffrey's hyper-prior over the Laplace prior. In
\citep{Tipping:01}, a gamma hyper-prior is used in place of the
second-level exponential as in the hierarchical interpretation of
the Laplace prior.

To encourage sparsity in SVM, two strategies have been used. The
first one is to replace the $L_2$-norm by an $L_1$-norm of the
weights \citep{Bennett:92,Zhu:04}. The second strategy is to
explicitly add a cardinality constraint on the weights. This will
lead to a hard non-convex optimization problem; thus relaxations
must be applied \citep{Chan:07}. Under the maximum entropy
discrimination models, feature selection was studied in
\citep{Jebara:00} by introducing a set of structural variables. It
is straightforward to generalize it to the structured learning case
but the resultant learning problem can be highly complex and
approximation must be developed.

Although the parameter distribution $p(\wv)$ in
Theorem~\ref{thrm_medn} has a similar form as that of the Bayesian
Conditional Random Fields (BCRFs) \citep{Qi:05}, MaxEnDNet is
fundamentally different from BCRFs as we have stated.
\cite{Dredze:08} present an interesting confidence-weighted linear
classification method, which automatically estimates the mean and
variance of model parameters in online learning. The procedure is
similar to (but indeed different from) our variational Bayesian
method of Laplace MaxEnDNet.

Finally, some of the results shown in this paper appeared in the
conference paper \citep{Zhu:08a}.

\section{Conclusions and Future Work}

To summarize, we have presented a general theory of maximum entropy
discrimination Markov networks for structured input/output learning
and prediction. This formalism offers a formal paradigm for
integrating both generative and discriminative principles and the
Bayesian regularization techniques for learning structured
prediction models. It subsumes popular methods such as support
vector machines, maximum entropy discrimination
models~\citep{Jaakkola:99}, and maximum margin Markov networks as
special cases, and therefore inherits all the merits of these
techniques.

The MaxEnDNet model offers a number of important advantage over
conventional structured prediction methods, including: 1) modeling
averaging, which leads to a PAC-Bayesian bound on generalization
error; 2) entropic regularization over max-margin learning, which
can be leveraged to learn structured prediction models that are
simultaneously primal and dual sparse; and 3) latent structures
underlying the structured input/output variables, which enables
better incorporation of domain knowledge in model design and
semi-supervised learning based on partially labeled data. In this
paper, we have discussed in detail the first two aspects, and the
third aspect is explored in~\citep{Zhu:08b}. We have also shown that
certain instantiations of the MaxEnDNet model, such as the LapMEDN
that achieves primal and dual sparsity, can be efficiently trained
based on an iterative optimization scheme that employs existing
techniques such as the variational Bayes approximation and the
convex optimization procedures that solve the standard M$^3$N. We
demonstrated that on synthetic data the LapMEDN can recover the
sparse model as well as the sparse $L_1$-regularized MAP estimation,
and on real data sets LapMEDN can achieve superior performance.

Overall, we believe that the MaxEnDNet model can be extremely
general and adaptive, and it offers a promising new framework for
building more flexible, generalizable, and large scale structured
prediction models that enjoy the benefits from both generative and
discriminative modeling principles. While exploring novel
instantiations of this model will be an interesting direction to
pursue, development of more efficient learning algorithms,
formulation of tighter but easy to solve convex relaxations, and
adapting this model to challenging applications such as statistical
machine translation, and structured associations of genome markers
to complex disease traits could also lead to fruitful results.

\section*{Acknowledgements}
This work was done while Jun Zhu was with the SAILING Lab directed
by Eric Xing at Carnegie Mellon University, under a visiting
scholarship sponsored by the National Natural Science Foundation of
China. The authors would like to thank Andrew Bagnell for sharing
their implementation of the sub-gradient algorithm, Ivor Tsang and
Andr$\acute{e}$ Martins for inspiring discussions, and all members
of the SAILING Lab for helpful discussions. Eric Xing is supported
by NSF grants CCF-0523757, DBI-0546594, IIS-0713379, DBI- 0640543,
and a Sloan Research Fellowship in Computer Science; Jun Zhu is
supported by the National Natural Science Foundation of China,
Grant. No. 60321002; the National Key Foundation R\&D Projects,
Grant No. 2004CB318108 and 2007CB311003; and Microsoft Fellowship.

\appendix

\section*{Appendix A. $L_1$-M$^3$N and its Lagrange-Dual}
Based on the $L_1$-norm regularized SVM \citep{Zhu:04,Bennett:92}, a straightforward formulation of $L_1$-M$^3$N is as follows, \\[-0.9cm]

\begin{eqnarray}
&&\min_{\wv, \xi}~\frac{1}{2} \Vert \wv \Vert + C \sum_{i=1}^N \xi_i \nonumber \\
&&\mathrm{s.t.}~~\wv^\top \Delta \fv_i(\yv) \geq \Delta \ell_i(\yv)
- \xi_i;~\xi_i \geq 0,~ \forall i,~\forall \yv \neq \yv^i \nonumber
\end{eqnarray}\\[-0.9cm]

\noindent where $\Vert . \Vert$ is the $L_1$-norm. $\Delta
\fv_i(\yv) = \fv(\xv^i, \yv^i) - \fv(\xv^i, \yv)$, and $\Delta
\ell_i(\yv)$ is a loss function. Another equivalent
formulation\footnote{See \citep{Taskar:06} for the transformation
technique.} is as follows:
\begin{eqnarray}
&&\min_{\wv, \xi}~ C \sum_{i=1}^N \xi_i \nonumber \\
&&\mathrm{s.t.}~~\left\{ \begin{array}{l}
    \Vert \wv \Vert \leq \lambda \\
    \wv^\top \Delta \fv_i(\yv) \geq \Delta
\ell_i(\yv) - \xi_i;~\xi_i \geq 0,~\forall i,~\forall \yv \neq \yv^i
    \end{array} \right. \nonumber
\end{eqnarray}\\[-0.9cm]

To derive the convex dual problem, we introduce a dual variable
$\alpha_i(\yv)$ for each constraint in the former formulation and
form the Lagrangian as follows,\\[-0.6cm]

\begin{displaymath} L(\alpha, \wv, \xi) = \frac{1}{2} \Vert
\wv \Vert + C \sum_{i=1}^N \xi_i - \sum_{i, \yv \neq \yv^i}
\alpha_i(\yv) \big( \wv^\top \Delta \fv_i(\yv) - \Delta \ell_i(\yv)
+ \xi_i \big).
\end{displaymath}

By definition, the Lagrangian dual is,\\[-0.8cm]

\setlength\arraycolsep{1pt}{\begin{eqnarray} L^\star(\alpha) && =
\inf_{\wv, \xi} L(\alpha, \wv, \xi) \nonumber \\
&& = \inf_{\wv} \Big \lbrack  \frac{1}{2} \Vert \wv \Vert - \sum_{i,
\yv \neq \yv^i} \alpha_i(\yv) \wv^\top \Delta \fv_i(\yv) \Big\rbrack
+ \inf_{\xi} \Big \lbrack C \sum_{i=1}^N \xi_i - \sum_{i,
\yv \neq \yv^i} \alpha_i(\yv) \xi_i \Big\rbrack + \ell \nonumber \\
&& = -\sup_{\wv} \Big \lbrack \wv^\top \big( \sum_{i, \yv \neq
\yv^i} \alpha_i(\yv) \Delta \fv_i(\yv) \big) - \frac{1}{2} \Vert \wv
\Vert \Big\rbrack - \sup_{\xi} \Big \lbrack \sum_{i, \yv \neq \yv^i}
\alpha_i(\yv) \xi_i - C \sum_{i=1}^N \xi_i \Big\rbrack + \ell,
\nonumber
\end{eqnarray}}where $\ell = \sum_{i,
\yv \neq \yv^i} \alpha_i(\yv) \Delta \ell_i(\yv)$.

Again, by definition, the first term on the right-hand side is the
convex conjugate of $\phi(\wv) = \frac{1}{2} \Vert \wv \Vert$ and
the second term is the conjugate of $U(\xi) = C \sum_{i=1}^N \xi_i$.
It is easy to show that,
\begin{displaymath} \phi^\star( \alpha ) = \indicator_{\D
\infty}( |\sum_{i, \yv \neq \yv^i} \alpha_i(\yv) \Delta
\fv_i^k(\yv)| \leq \frac{1}{2},~\forall 1 \leq k \leq K),
\end{displaymath}and
\begin{displaymath} U^\star(\alpha) = \indicator_{\D
\infty}(\sum_{\yv \neq \yv^i} \alpha_i(\yv) \leq C,~\forall i),
\end{displaymath}
\noindent where as defined before $\indicator_{\D \infty}(\cdot)$ is
an indicator function that equals zero when its argument is true and
infinity otherwise. $\Delta \fv_i^k(\yv) = f_k(\xv^i, \yv^i) -
f_k(\xv^i, \yv)$.

Therefore, we get the dual problem as follows,\\[-1cm]

\begin{eqnarray}
&&\max_{\alpha}~~\sum_{i, \yv \neq \yv^i} \alpha_i(\yv) \Delta \ell_i(\yv) \nonumber \\
&& \textrm{s.t.}~~\vert \sum_{i, \yv \neq \yv^i} \alpha_i(\yv)
\Delta \fv_i^k(\yv) \vert \leq \frac{1}{2},~\forall k \nonumber \\
&&~~~~~\sum_{\yv \neq \yv^i} \alpha_i(\yv) \leq C,~\forall i.
\nonumber
\end{eqnarray}

\section*{Appendix B. Proofs of Theorems and Corollaries}

\subsection*{Appendix B.1. Proof of Theorem \ref{thrm_medn}}

\begin{proof}
As we have stated, P1 is a convex program and satisfies the Slater's
condition. To compute its convex dual, we introduce a non-negative
dual variable $\alpha_i(\yv)$ for each constraint in $\mathcal{F}_1$
and another non-negative dual variable $c$ for the normalization
constraint $\int p(\wv) \, \ud \wv \, = 1$. This gives rise to the
following Lagrangian: \setlength\arraycolsep{1pt} {\begin{eqnarray}
L(p(\wv), \xi, \alpha, c) && = KL(p(\wv) || p_0(\wv) ) + U(\xi) + c ( \int p(\wv) \, \ud \wv \, - 1 ) \nonumber \\
&& - \sum_{i, \yv \neq \yv^i} \alpha_i(\yv) \big( \int p(\wv)
\lbrack \Delta F_i(\yv; \wv) - \Delta \ell_i(\yv) \rbrack \, \ud \wv
\, + \xi_i \big) . \nonumber
\end{eqnarray}} \\[-0.9cm]

\noindent The Lagrangian dual function is defined as
$L^\star(\alpha, c) \triangleq \inf_{p(\wv); \xi} L(p(\wv), \xi,
\alpha, c)$. Taking the derivative of $L$ w.r.t $p(\wv)$, we get,
\begin{displaymath} \frac{\partial{L}}{\partial{p(\wv)}} = 1 + c +
\log \frac{p(\wv)}{p_0(\wv)} - \sum_{i, \yv \neq \yv^i}
\alpha_i(\yv) \lbrack \Delta F_i(\yv; \wv) - \Delta \ell_i(\yv)
\rbrack.
\end{displaymath}
Setting the derivative to zero, we get the following expression of
distribution $p(\wv)$,
\begin{displaymath} p(\wv) = \frac{1}{Z(\alpha)} p_0(\wv)
\exp \big \lbrace \sum_{i, \yv \neq \yv^i} \alpha_i(\yv) \lbrack
\Delta F_{i}(\yv; \wv) - \Delta \ell_{i}(\yv) \rbrack \big \rbrace ,
\end{displaymath}
\noindent where $Z(\alpha) \triangleq \int p_0(\wv) \exp \big
\lbrace \sum_{i, \yv \neq \yv^i} \alpha_i(\yv) \lbrack \Delta
F_{i}(\yv; \wv) - \Delta \ell_{i}(\yv) \rbrack \big \rbrace \, \ud
\wv \,$ is a normalization constant and $c = -1 + \log Z(\alpha)$.

Substituting $p(\wv)$ into $L^\star$, we obtain,
\setlength\arraycolsep{1pt} {\begin{eqnarray} L^\star(\alpha, c) &=&
\inf_{p(\wv); \xi} \big( -\log Z(\alpha) + U(\xi) - \sum_{i, \yv
\neq \yv^i} \alpha_i(\yv) \xi_i \big) \nonumber \\
&=& -\log Z(\alpha) + \inf_{\xi} \big( U(\xi) - \sum_{i,
\yv \neq \yv^i} \alpha_i(\yv) \xi_i \big) \nonumber \\
&=& -\log Z(\alpha) - \sup_{\xi} \big( \sum_{i, \yv \neq \yv^i}
\alpha_i(\yv) \xi_i - U(\xi) \big) \nonumber \\
&=& - \log Z(\alpha) - U^\star(\alphav), \nonumber
\end{eqnarray}}\noindent which is the objective in the dual problem D1. 
The $\{\alpha_i(\yv)\}$ derived from D1 lead to the optimum $p(\wv)$
according to Eq. (\ref{wdist}).
\end{proof}

\subsection*{Appendix B.2. Proof of Theorem \ref{thrm_gaussianmedn}}

\begin{proof}
Replacing $p_0(\wv)$ and $\Delta F_i(\yv; \wv)$ in Eq. (\ref{wdist})
with $\mathcal{N}(\wv |0, I)$ and $\wv^\top \Delta \fv_i(\yv)$
respectively, we can obtain the following closed-form expression of
the $Z(\alpha)$ in $p(\wv)$:\\[-0.9cm]

\setlength\arraycolsep{1pt}{\begin{eqnarray} Z(\alpha) && \triangleq
\int \mathcal{N}(\wv |0, I) \exp \Big \lbrace \sum_{i, \yv \neq
\yv^i} \alpha_i(\yv) \lbrack \wv^\top \Delta
\fv_i(\yv) - \Delta \ell_{i}(\yv) \rbrack \Big \rbrace \, \ud \wv \, \nonumber \\
&& = \int (2\pi)^{-\frac{K}{2}} \exp \Big \lbrace -\frac{1}{2}
\wv^\top \wv + \sum_{i, \yv \neq \yv^i} \alpha_i(\yv) \lbrack
\wv^\top \Delta \fv_{i}(\yv) -
\Delta \ell_{i}(\yv) \rbrack \Big \rbrace \, \ud \wv \, \nonumber \\
&& = \exp \Big( - \sum_{i, \yv \neq \yv^i} \alpha_i(\yv) \Delta
\ell_{i}(\yv) + \frac{1}{2} \Vert \sum_{i,\yv \neq \yv^i}
\alpha_i(\yv) \Delta \fv_{i}(\yv) \Vert^2 \Big). \nonumber
\end{eqnarray}}\\[-0.9cm]

Substituting the normalization factor into the general dual problem
D1, we get the dual problem of Gaussian MaxEnDNet. As we have
stated, the constraints $\sum_{\yv \neq \yv^i} \alpha_i(\yv) = C$
are due to the conjugate of $U(\xi) = C \sum_i \xi_i$.

For prediction, again replacing $p_0(\wv)$ and $\Delta F_i(\yv;
\wv)$ in Eq. (\ref{wdist}) with $\mathcal{N}(\wv |0, I)$ and
$\wv^\top \Delta \fv_i(\yv)$ respectively, we can get $p(\wv) =
\mathcal{N}(\wv|\mu, I)$, where $\mu = \sum_{i,\yv \neq \yv^i}
\alpha_i(\yv) \Delta \fv_{i}(\yv)$. Substituting $p(\wv)$ into the
predictive function $h_1$, we can get $h_1(\xv) = \arg \max_{\yv \in
\mathcal{Y}(\xv)} \mu^\top \fv(\xv, \yv) = (\sum_{i,\yv \neq \yv^i}
\alpha_i(\yv) \Delta \fv_{i}(\yv))^\top \fv(\xv, \yv)$, which is
identical to the prediction rule of the standard M$^3$N
\citep{Taskar:03} because the dual parameters are achieved by
solving the same dual problem.
\end{proof}

\subsection*{Appendix B.3. Proof of Corollary \ref{cor_primal_gaussianmedn}}

\begin{proof}
Suppose $(p^\star(\wv), \xi^\star)$ is the optimal solution of P1,
then we have: for any $(p(\wv), \xi),~p(\wv) \in \mathcal{F}_1$ and
$\xi \geq 0$, $$KL(p^\star(\wv) || p_0(\wv)) + U(\xi^\star) \leq
KL(p(\wv) || p_0(\wv)) + U(\xi).$$

From Theorem~\ref{thrm_gaussianmedn}, we conclude that the optimum
predictive parameter distribution is $p^\star(\wv)=\mathcal{N}(\wv |
\mu^\star, I)$. Since $p_0(\wv)$ is also normal, for any
distribution $p(\wv)=\mathcal{N}(\wv | \mu, I)$\footnote{Although
$\mathcal{F}_1$ is much richer than the set of normal distributions
with an identity covariance matrix, Theorem~\ref{thrm_gaussianmedn}
shows that the solution is a restricted normal distribution. Thus,
it suffices to consider only these normal distributions in order to
learn the mean of the optimum distribution. The similar argument
applies to the proof of Corollary~\ref{cor_primal_lapmaxendnet}. }
with several steps of algebra it is easy to show that $KL(p(\wv) |
p_0(\wv) ) = \frac{1}{2} \mu^\top \mu $. Thus, we can get: for any
$(\mu, \xi),~\mu \in \lbrace \mu:~\mu^\top \Delta \fv_i(\yv) \geq
\Delta \ell_{i}(\yv) - \xi_i,~\forall i,~\forall \yv \neq \yv^i
\rbrace$ and $\xi \geq 0 $,
$$\frac{1}{2} (\mu^\star)^\top (\mu^\star) + U(\xi^\star) \leq
\frac{1}{2} \mu^\top \mu + U(\xi^\star),$$ which means the mean of
the optimum posterior distribution under a Gaussian MaxEnDNet is
achieved by solving a primal problem as stated in the Corollary.
\end{proof}

\subsection*{Appendix B.4. Proof of Corollary \ref{cor_primal_lapmaxendnet}}

\begin{proof}
The proof follows the same structure as the above proof of
Corollary~\ref{cor_primal_gaussianmedn}. Here, we only present the
derivation of the KL-divergence under the Laplace MaxEnDNet.

Theorem~\ref{thrm_medn} shows that the general posterior
distribution is~$p(\wv) = \frac{1}{Z(\alpha)} p_0(\wv) \exp(
\wv^\top \eta - \sum_{i,\yv \neq \yv^i} \alpha_i(\yv) \Delta
\ell_i(\yv))$ and $Z(\alpha) = \exp(-\sum_{i,\yv \neq \yv^i}
\alpha_i(\yv) \Delta \ell_i(\yv)) \prod_{k=1}^K
\frac{\lambda}{\lambda - \eta_k^2}$ for the Laplace MaxEnDNet as
shown in Eq. (\ref{normfactor_lapmedn}). Use the definition of
KL-divergence and we can get:
\begin{eqnarray}
KL(p(\wv) | p_0(\wv) ) = \langle \wv \rangle_{p}^\top \eta -
\sum_{k=1}^K \log \frac{\lambda}{\lambda - \eta_k^2} = \sum_{k=1}^K
\mu_k  \eta_k - \sum_{k=1}^K \log \frac{\lambda}{\lambda -
\eta_k^2}, \nonumber
\end{eqnarray}

Corollary \ref{remark_postshrinkage} shows that $\mu_k =
\frac{2\eta_k}{\lambda - \eta_k^2},~\forall 1 \leq k \leq K$. Thus,
we get $\frac{\lambda}{\lambda - \eta_k^2} = \frac{\lambda
\mu_k}{2\eta_k}$ and a set of equations:~$\mu_k \eta_k^2 + 2\eta_k -
\lambda \mu_k = 0,~\forall 1 \leq k \leq K$. To solve these
equations, we consider two cases. First, if $\mu_k = 0$, then
$\eta_k = 0$. Second, if $\mu_k \neq 0$, then we can solve the
quadratic equation to get $\eta_k$:~$\eta_k = \frac{-1 \pm \sqrt{1 +
\lambda \mu_k^2} } {\mu_k}$. The second solution includes the first
one since we can show that when $\mu_k \to 0$, $\frac{-1 \pm \sqrt{1
+ \lambda \mu_k^2} } {\mu_k} \to 0$ by using the {\it L'Hospital's
Rule}. Thus, we get:
\begin{eqnarray}
\mu_k \eta_k = -1 \pm \sqrt{ \lambda \mu_k^2 + 1}. \nonumber
\end{eqnarray}

Since $\eta_k^2 < \lambda$ (otherwise the problem is not bounded),
$\mu_k \eta_k$ is always positive. Thus, only the solution $\mu_k
\eta_k = -1 + \sqrt{ 1 + \lambda \mu_k^2}$~is feasible. So, we get:
$$ \frac{\lambda}{\lambda - \eta_k^2} = \frac{\lambda
\mu_k^2}{2(\sqrt{\lambda \mu_k^2+1}-1)} = \frac{\sqrt{\lambda
\mu_k^2+1} + 1}{2},$$ and
\setlength\arraycolsep{1pt}{\begin{eqnarray} KL(p(\wv) | p_0(\wv))
&=& \sum_{k=1}^K \Big( \sqrt{ \lambda \mu_k^2 + 1} - \log
\frac{\sqrt{\lambda \mu_k^2+1} + 1}{2}\Big) - K \nonumber \\
&=& \sqrt{\lambda} \sum_{k=1}^K \Big( \sqrt{ \mu_k^2 +
\frac{1}{\lambda}} - \frac{1}{\sqrt{\lambda}} \log
\frac{\sqrt{\lambda \mu_k^2+1} + 1}{2}\Big) - K. \nonumber
\end{eqnarray}}

\noindent Applying the same arguments as in the above proof of
Corollary~\ref{cor_primal_gaussianmedn} and using the above result
of the KL-divergence, we get the problem in
Corollary~\ref{cor_primal_lapmaxendnet}, where the constant $-K$ is
ignored. The margin constraints defined with the mean $\mu$ are due
to the linearity assumption of the discriminant functions.
\end{proof}

\subsection*{Appendix B.5. Proof of Theorem \ref{thrm_pacbayes}}

We follow the same structure as the proof of PAC-Bayes bound for
binary classifier \citep{Langford:01} and employ the similar
technique to generalize to multi-class problems as in
\citep{Schapire:98}. Recall that the output space is $\mathcal{Y}$,
and the base discriminant function is $F(\ \cdot \ ; \wv) \in
\mathcal{H}: \mathcal{X} \times \mathcal{Y} \to \lbrack -c,
c\rbrack$, where $c  > 0$ is a constant. Our averaging model is
specified by $h(\xv, \yv) = \langle F(\xv, \yv; \wv)
\rangle_{p(\wv)}$. We define the margin of an example $(\xv,
\yv)$ for such a function $h$ as, \\[-0.9cm]

\begin{equation}
M(h, \xv, \yv) = h(\xv, \yv) - \max_{\yv^\prime \neq \yv} h(\xv, \yv^\prime).
\label{mrg}
\end{equation}\\[-0.9cm]

\noindent Thus, the model $h$ makes a wrong prediction on $(\xv,
\yv)$ only if $M(h, \xv, \yv) \leq 0$. Let $Q$ be a distribution
over $\mathcal{X} \times \mathcal{Y}$, and let $\mathcal{D}$ be a
sample of $N$ examples independently and randomly drawn from $Q$.
With these definitions, we have the PAC-Bayes theorem. For easy
reading, we copy the theorem in the following:

{\bf Theorem~\ref{thrm_pacbayes} (PAC-Bayes Bound of MaxEnDNet)}
{\it Let $p_0$ be any continuous probability distribution over
$\mathcal{H}$ and let $\delta \in (0, 1)$. If $F(\ \cdot \ ; \wv)
\in \mathcal{H}$ is bounded by $\pm c$ as above, then with
probability at least $1 - \delta$, for a random sample $\mathcal{D}$
of $N$ instances from $Q$, for every distribution $p$ over
$\mathcal{H}$, and for all margin thresholds
$\gamma > 0$:\\[-0.9cm]

\setlength\arraycolsep{1pt} {\begin{eqnarray} \mathrm{Pr}_{Q} ( M(h,
\xv, \yv ) \leq 0 ) \leq \mathrm{Pr}_{\mathcal{D}} ( M(h, \xv, \yv)
\leq \gamma ) + O \Big( \sqrt{ \frac{\gamma^{-2} KL(p||p_0) \ln (N
|\mathcal{Y}|) + \ln N + \ln \delta^{-1}}{N} } \Big), \nonumber
\end{eqnarray}}\\[-0.9cm]

\noindent where $\mathrm{Pr}_{Q}(.)$ and
$\mathrm{Pr}_{\mathcal{D}}(.)$ represent the probabilities of events
over the true distribution $Q$, and over the empirical distribution
of $\mathcal{D}$, respectively.}

\begin{proof} Let $m$ be any natural number. For every distribution $p$, we independently draw $m$ base models (i.e., discriminant functions) $F_i \sim p$ at random. We also independently draw $m$ variables $\mu_i \sim U(\lbrack -c, c\rbrack)$, where $U$ denote the uniform distribution. We define the binary functions $g_i : \mathcal{X} \times \mathcal{Y} \to \lbrace -c,
+c \rbrace$ by: \\[-0.6cm]

\begin{displaymath}
g_i(\xv, \yv; F_i, \mu_i) = 2c I(\mu_i < F_i(\xv, \yv)) - c.
\end{displaymath} \\[-1cm]

With the $F_i$, $\mu_i$, and $g_i$, we define $\mathcal{H}_m$ as,\\[-0.6cm]

\begin{displaymath}
\mathcal{H}_m = \big\lbrace f: (\xv, \yv) \mapsto \frac{1}{m} \sum_{i=1}^m g_i(\xv, \yv; F_i, \mu_i) | F_i \in \mathcal{H}, \mu_i \in \lbrack -c, c \rbrack \big\rbrace.
\end{displaymath} \\[-0.8cm]

We denote the distribution of $f$ over the set $\mathcal{H}_m$ by $p^m$. For a fixed pair $(\xv, \yv)$, the quantities $g_i(\xv, \yv; F_i, \mu_i)$ are i.i.d bounded random variables with the mean:\\[-1cm]

\setlength\arraycolsep{1pt}{\begin{eqnarray} \langle g_i ( \xv, \yv;
F_i, \mu_i ) \rangle_{F_i \sim p, \mu_i \sim U[-c, c]}
&&= \langle (+c)p\lbrack \mu_i \leq F_i(\xv, \yv) | F_i \rbrack + (-c) p\lbrack \mu_i > F_i(\xv, \yv) | F_i \rbrack \rangle_{F_i \sim p} \nonumber \\
&&= \langle \frac{1}{2c} c ( c + F_i(\xv, \yv) ) - \frac{1}{2c} c ( c - F_i(\xv, \yv) ) \rangle_{F_i \sim p} \nonumber \\
&&= h(\xv, \yv). \nonumber
\end{eqnarray}}\\[-1cm]

Therefore, $\langle f(\xv, \yv) \rangle_{f \sim p^m} = h(\xv, \yv)$.
Since $f(\xv, \yv)$ is the average over $m$ i.i.d bounded variables,
Hoeffding's inequality applies. Thus, for
every $(\xv, \yv)$,\\[-0.6cm]

\begin{displaymath}
\mathrm{Pr}_{f \sim p^m} \lbrack f(\xv, \yv) - h(\xv, \yv) > \xi \rbrack \leq e^{-\frac{m}{2c^2}
\xi^2}.
\end{displaymath}\\[-0.8cm]

For any two events $A$ and $B$, we have the inequality,\\[-0.6cm]

\begin{displaymath}
\mathrm{Pr}(A) = \mathrm{Pr}(A, B) + \mathrm{Pr}(A, \bar{B}) \leq \mathrm{Pr}(B) + \mathrm{Pr}(\bar{B} | A).
\end{displaymath}\\[-1cm]

Thus, for any $\gamma > 0$ we have \\[-0.9cm]

\setlength\arraycolsep{1pt}{\begin{eqnarray} \mathrm{Pr}_{Q}
\big\lbrack M(h, \xv, \yv) \leq 0 \big\rbrack \leq \mathrm{Pr}_{Q}
\big\lbrack M(f, \xv, \yv) \leq \frac{\gamma}{2} \big\rbrack +
\mathrm{Pr}_{Q} \big\lbrack M(f, \xv, \yv) > \frac{\gamma}{2} | M(h,
\xv, \yv) \leq 0 \big\rbrack. \label{event_ineqn1}
\end{eqnarray}}\\[-0.8cm]

Fix $h, \xv,$ and $\yv$, and let $\yv^\prime$ achieve the margin in (\ref{mrg}). Then, we get\\[-0.6cm]

\begin{displaymath}
M(h, \xv, \yv) = h(\xv, \yv) - h(\xv, \yv^\prime),~\textrm{and}~M(f, \xv, \yv) \leq f(\xv, \yv) - f(\xv, \yv^\prime).
\end{displaymath}\\[-0.9cm]

\noindent With these two results, since $\langle f(\xv, \yv) - f(\xv, \yv^\prime) \rangle_{f \sim p^m} =
h(\xv, \yv) - h(\xv, \yv^\prime)$, we can get \\[-1.0cm]

\setlength\arraycolsep{1pt}{\begin{eqnarray} \mathrm{Pr}_{Q}
\big\lbrack  M(f, \xv, \yv) > \frac{\gamma}{2} | M(h, \xv, \yv) \leq
0 \big\rbrack &&\leq \mathrm{Pr}_{Q} \big\lbrack f(\xv, \yv) -
f(\xv, \yv^\prime) > \frac{\gamma}{2} | M(h,
\xv, \yv) \leq 0 \big\rbrack \nonumber \\
&& \leq \mathrm{Pr}_{Q} \big\lbrack f(\xv, \yv) - f(\xv, \yv^\prime) - M(h,
\xv, \yv) > \frac{\gamma}{2} \big\rbrack \nonumber \\
&& \leq e^{- \frac{m \gamma^2}{32c^2} },
\label{hoeffding1}
\end{eqnarray}}\\[-1.0cm]

\noindent where the first two inequalities are due to the fact that
if two events $A \subseteq B$, then $p(A) \leq p(B)$, and the last
inequality is due to the Hoeffding's inequality.

Substitute (\ref{hoeffding1}) into (\ref{event_ineqn1}), and we get,\\[-0.7cm]

\begin{displaymath} \mathrm{Pr}_{Q}
\big\lbrack M(h, \xv, \yv) \leq 0 \big\rbrack \leq \mathrm{Pr}_{Q}
\big\lbrack M(f, \xv, \yv) \leq \frac{\gamma}{2}  \big\rbrack  +
e^{- \frac{ m \gamma^2}{32c^2} },
\end{displaymath}\\[-0.9cm]

\noindent of which the left hand side does not depend on $f$.
We take the expectation over $f \sim p^m$ on both sides and get, \\[-0.9cm]

\setlength\arraycolsep{1pt}{\begin{eqnarray} \mathrm{Pr}_{Q}
\big\lbrack M(h, \xv, \yv) \leq 0 \big\rbrack \leq \langle
\mathrm{Pr}_{Q} \big\lbrack M(f, \xv, \yv) \leq \frac{\gamma}{2}
\big\rbrack \rangle_{f \sim p^m} + e^{- \frac{ m \gamma^2}{32c^2} }.
\label{exp_ineqn}
\end{eqnarray}}\\[-0.9cm]

\indent Let $p_0^m$ be a prior distribution on $\mathcal{H}_m$.
$p_0^m$ is constructed from $p_0$ over $\mathcal{H}$ exactly as
$p^m$ is constructed from $p$. Then, $KL( p^m || p_0^m ) = m KL( p
|| p_0)$. By the PAC-Bayes theorem \citep{McAllester:99}, with
probability at least $1 - \delta$ over sample $\mathcal{D}$, the following bound holds for any distribution $p$, \\[-0.9cm]

\setlength\arraycolsep{1pt}{\begin{eqnarray} \langle \mathrm{Pr}_{Q}
\big\lbrack M(f, \xv, \yv) \leq \frac{\gamma}{2} \big\rbrack
\rangle_{f \sim p^m}
&& \leq \langle \mathrm{Pr}_{\mathcal{D}} \big\lbrack M(f, \xv, \yv) \leq \frac{\gamma}{2}  \big\rbrack \rangle_{f \sim p^m} \nonumber \\
&&+ \sqrt{ \frac{m KL(p || p_0) + \ln N + \ln \delta^{-1} + 2}{2N - 1}}.
\label{pac}
\end{eqnarray}}\\[-0.9cm]

By the similar statement as in (\ref{event_ineqn1}), for every $f \in \mathcal{H}_m$ we have, \\[-0.9cm]

\setlength\arraycolsep{1pt}{\begin{eqnarray}
\mathrm{Pr}_{\mathcal{D}} \big\lbrack M(f, \xv, \yv) \leq
\frac{\gamma}{2} \big\rbrack \leq \mathrm{Pr}_{\mathcal{D}}
\big\lbrack M(h, \xv, \yv) \leq \gamma \big\rbrack +
\mathrm{Pr}_{\mathcal{D}} \big\lbrack M(f, \xv, \yv) \leq
\frac{\gamma}{2} | M(h, \xv, \yv) > \gamma \big\rbrack.
\label{event_ineqn2}
\end{eqnarray}}\\[-0.9cm]

Rewriting the second term on the right-hand side of (\ref{event_ineqn2}), we get\\[-0.9cm]

\begin{eqnarray}
\mathrm{Pr}_{\mathcal{D}} \big\lbrack M(f, \xv, \yv) \leq \frac{\gamma}{2} | M(h, \xv, \yv) > \gamma \big\rbrack
&& = \mathrm{Pr}_{\mathcal{D}} \big \lbrack \exists \yv^\prime \neq \yv: \Delta f(\xv, \yv^\prime) \leq \frac{\gamma}{2} | \forall
\yv^\prime \neq \mathrm{y}: \Delta h(\xv, \yv^\prime) > \gamma \big\rbrack \nonumber \\
&& \leq \mathrm{Pr}_{\mathcal{D}} \big \lbrack \exists \yv^\prime \neq \yv: \Delta f(\xv, \yv^\prime) \leq \frac{\gamma}{2} | \Delta h(\xv, \yv^\prime) > \gamma \big\rbrack \nonumber \\
&& \leq \sum_{\yv^\prime \neq \yv} \mathrm{Pr}_{\mathcal{D}} \big \lbrack \Delta f(\xv, \yv^\prime) \leq \frac{\gamma}{2} | \Delta h(\xv,
\yv^\prime) > \gamma \big\rbrack \nonumber \\
&& \leq (|\mathcal{Y}| - 1) e^{- \frac{m \gamma^2}{32c^2}},
\label{hoeffding2}
\end{eqnarray}\\[-0.9cm]

\noindent where we use $\Delta f(\xv, \yv^\prime)$ to denote $f(\xv,
\yv) - f(\xv, \yv^\prime)$, and use $\Delta h(\xv, \yv^\prime)$ to
denote $h(\xv, \yv) - h(\xv, \yv^\prime)$.

Put (\ref{exp_ineqn}), (\ref{pac}), (\ref{event_ineqn2}), and (\ref{hoeffding2}) together, then we get following bound holding for any fixed $m$ and $\gamma > 0$,\\[-0.9cm]

\setlength\arraycolsep{1pt}{\begin{eqnarray} \mathrm{Pr}_{Q} \big
\lbrack M(h, \xv, \yv) \leq 0 \big\rbrack && \leq
\mathrm{Pr}_{\mathcal{D}} \big \lbrack M(h, \xv, \yv) \leq \gamma
\big\rbrack + |\mathcal{Y}|e^{-\frac{m \gamma^2}{32c^2}} + \sqrt{
\frac{m KL(p || p_0) + \ln N + \ln \delta^{-1} + 2} {2N -1
}}.\nonumber
\end{eqnarray}} \\[-0.9cm]

To finish the proof, we need to remove the dependence on $m$ and
$\gamma$. This can be done by applying the union bound. By the
definition of $f$, it is obvious that if $f \in \mathcal{H}_m$ then
$f(\xv, \yv) \in \lbrace (2k-m)c/m: k = 0, 1, \dots, m \rbrace$.
Thus, even though $\gamma$ can be any positive value, there are no
more than $m+1$ events of the form $\lbrace M(f, \xv, \yv) \leq
\gamma / 2 \rbrace$. Since only the application of PAC-Bayes theorem
in (\ref{pac}) depends on $(m, \gamma)$ and all the other steps are
true with probability one, we just need to consider the union of
countably many events. Let $\delta_{m, k} = \delta / (m (m+1)^2)$,
then the union of all the possible events has a probability at most
$\sum_{m,k} \delta_{m, k} = \sum_m (m+1) \delta / (m(m+1)^2) =
\delta$. Therefore, with probability at least $1 - \delta$ over
random samples of $\mathcal{D}$, the following bound holds for all
$m$ and all $\gamma > 0$,\\[-0.9cm]

\setlength\arraycolsep{1pt}{\begin{eqnarray} \mathrm{Pr}_{Q} \big
\lbrack M(h, \xv, \yv) \leq 0 \big\rbrack -
\mathrm{Pr}_{\mathcal{D}} \big \lbrack M(h, \xv, \yv) \leq \gamma
\big\rbrack && \leq |\mathcal{Y}|e^{-\frac{m \gamma^2}{32c^2}} +
\sqrt{ \frac{m KL(p || p_0) + \ln N + \ln \delta_{m,k}^{-1} + 2} {2N
-1 }}
\nonumber \\
&&\leq |\mathcal{Y}|e^{-\frac{m \gamma^2}{32c^2}} + \sqrt{ \frac{m KL(p || p_0) + \ln N + 3 \ln \frac{m+1}{\delta} +
2} {2N -1 }} \nonumber
\end{eqnarray}}\\[-0.9cm]

Setting $m = \lceil 16 c^2 \gamma^{-2} \ln \frac{N
|\mathcal{Y}|^2}{KL(p||p_0) + 1} \rceil$ gives the results in the
theorem.
\end{proof}


\bibliography{SparseM3N}
\end{document}